\documentclass[journal, final]{IEEEtran}

\ifCLASSINFOpdf
\else
\fi
\usepackage{epsfig}
\usepackage{graphicx}
\usepackage{amsmath}
\usepackage{amssymb}
\usepackage{multirow, makecell}
\usepackage{lineno}
\usepackage{enumerate}
\usepackage{amsfonts}
\usepackage{gensymb}
\usepackage{bm}
\usepackage{diagbox}
\usepackage{tabularx}
\usepackage{rotating}
\usepackage[noend,ruled]{algorithm2e}
\newcolumntype{L}[1]{>{\raggedright\let\newline\\\arraybackslash\hspace{0pt}}m{#1}}
\newcolumntype{C}[1]{>{\centering\arraybackslash}m{#1}}
\newcolumntype{R}[1]{>{\raggedleft\let\newline\\\arraybackslash\hspace{0pt}}m{#1}}

\usepackage{xmpmulti}

\begin{document}
%
\title{SRG: Snippet Relatedness-based\\ Temporal Action Proposal Generator}
%
%
%
\author{Hyunjun~Eun,~\IEEEmembership{Student Member,~IEEE}, Sumin~Lee,~\IEEEmembership{Student Member,~IEEE}, \\ Jinyoung~Moon, Jongyoul~Park, Chanho~Jung, and
        Changick~Kim,~\IEEEmembership{Senior Member,~IEEE}
\thanks{Manuscript received August 1, 2019; revised October 9, 2019 and October
31, 2019; accepted November 6, 2019. This work was supported by Institute of Information \& Communications Technology Planning \& Evaluation (IITP) grant funded by the Korea government (MSIT) (No.B0101-15-0266, Development of High Performance Visual BigData Discovery Platform for Large-Scale Realtime Data Analysis). This article was recommended by Associate Editor H. Li. (Corresponding author:
Chanho Jung)}

\thanks{
H.~Eun, S.~Lee, and C.~Kim are with the School of Electrical Engineering,
Korea Advanced Institute of Science and Technology (KAIST), Daejeon 34141, South Korea
(e-mail: hj.eun@kaist.ac.kr; shum$\_$ming@kaist.ac.kr; changick@kaist.ac.kr).}
\thanks{J.~Moon and J.~Park are with the Electronics and Telecommunications Research Institute (ETRI), Daejeon 34129, South Korea
(e-mail: jymoon@etri.re.kr; jongyoul@etri.re.kr).}
\thanks{C.~Jung (corresponding author) is with the Department of Electrical Engineering, Hanbat National University, Daejeon 34158, South Korea
(e-mail: peterjung@hanbat.ac.kr).}

\thanks{Color versions of one or more of the figures in this article are available
online at http://ieeexplore.ieee.org.}
\thanks{Digital Object Identifier 10.1109/TCSVT.2019.2953187}
}

%
%

\markboth{ACCEPTED TO IEEE Transactions on Circuits and Systems for Video Technology}%
{}
%

\IEEEoverridecommandlockouts
\IEEEpubid{\begin{minipage}{\textwidth}\ \\[12pt]
\begin{center}
1051--8215~\copyright~2019 IEEE. Personal use of this material is permitted. However, permission to use this material\\ 
for any other purposes must be obtained from the IEEE by sending and email to pubs-permissions@ieee.org.
\end{center}
\end{minipage}}


\maketitle

\begin{abstract}
Recent temporal action proposal generation approaches have suggested integrating segment- and snippet score-based methodologies to produce proposals with high recall and accurate boundaries.
In this paper, different from such a hybrid strategy, we focus on the potential of the snippet score-based approach.
Specifically, we propose a new snippet score-based method, named Snippet Relatedness-based Generator (SRG), with a novel concept of ``snippet relatedness''.
Snippet relatedness represents which snippets are related to a specific action instance.
To effectively learn this snippet relatedness, we present ``pyramid non-local operations'' for locally and globally capturing long-range dependencies among snippets.
By employing these components, SRG first produces a 2D relatedness score map that enables the generation of various temporal intervals reliably covering most action instances with high overlap.
Then, SRG evaluates the action confidence scores of these temporal intervals and refines their boundaries to obtain temporal action proposals.
On THUMOS-14 and ActivityNet-1.3 datasets, SRG outperforms state-of-the-art methods for temporal action proposal generation.
Furthermore, compared to competing proposal generators, SRG leads to significant improvements in temporal action detection.
\end{abstract}


%
\IEEEpeerreviewmaketitle

\section{Introduction}
%
%
%
%
\IEEEPARstart{P}{eople} can easily share and watch large numbers of videos due to the growth of digital media platforms.
The analysis of these real-world videos has attracted considerable attention in the computer vision community.
One of the major tasks is temporal action detection, which aims to detect action instances in a long untrimmed video.
Specifically, the detection process includes localizing temporal boundaries and classifying the class of each instance.
Temporal action detection can be used in many applications such as smart surveillance \cite{iwashita2013bmvc, shu2015cvpr}, video summarization \cite{yao2015iccv, yao2016cvpr, zhang2016eccv}, and video retrieval \cite{chou2015tmm, douze2016ijcv, song2018pr}.

Similar to object proposals for object detection \cite{girshick2014cvpr, lin2017iccv, ren2015nips}, temporal action proposals play an important role in temporal action detection \cite{gao2017iccv, helibron2016cvpr, shou2016cvpr}.
Specifically, many methods \cite{chao2018cvpr, helibron2017cvpr, yuan2017cvpr, zhao2017iccv} for temporal action detection perform proposal generation followed by action classification.
For this reason, the performance of temporal action detection can be improved by generating a few proposals that capture temporal action instances with high overlap.

\begin{figure}[t!]
\centering{\includegraphics[width=.99\linewidth]{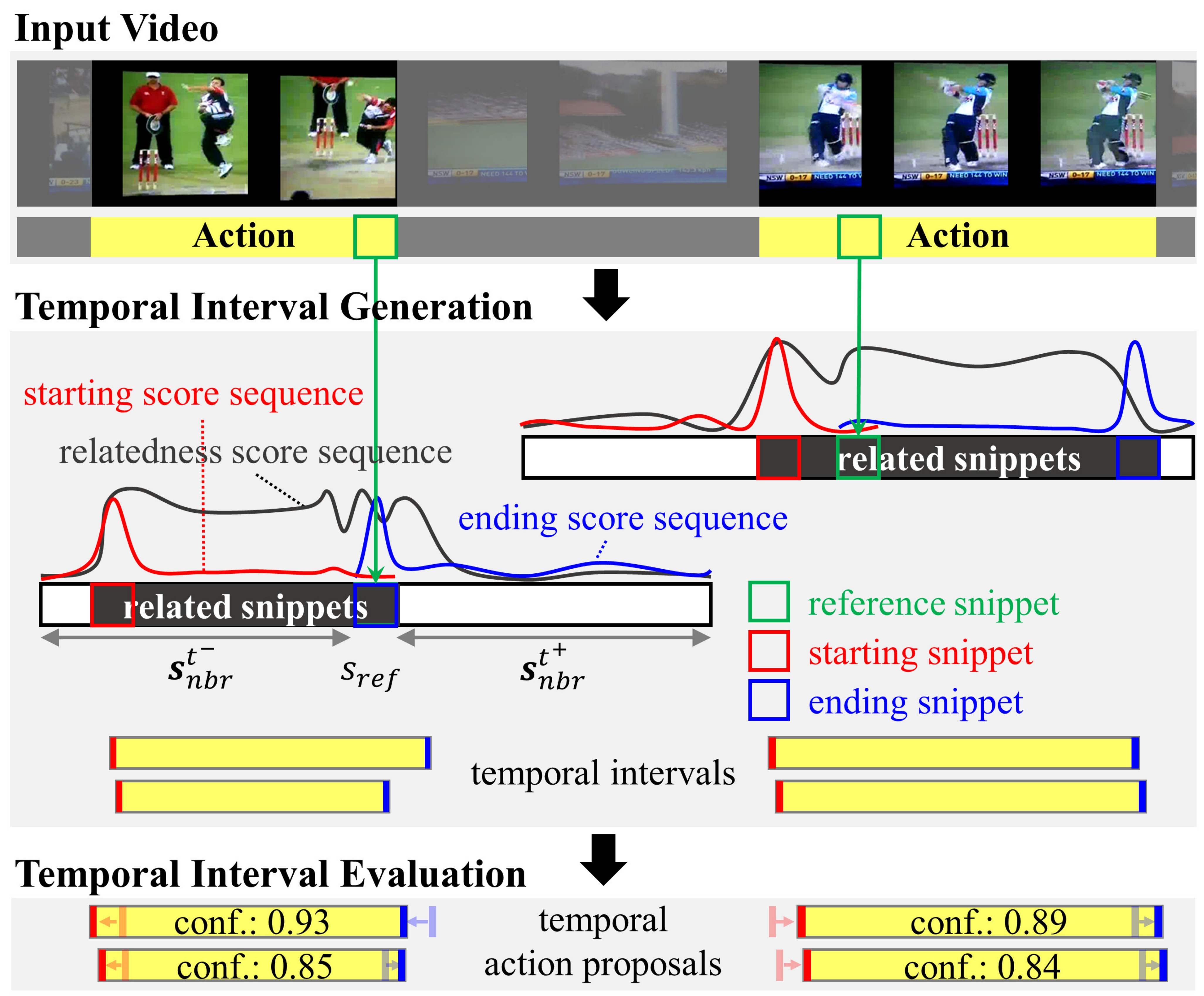}} \\
\caption{\label{fig1}Overview of our approach.
Given an input video, we first produce three types of score maps (i.e., relatedness, starting, and ending).
Each score map contains multiple score sequences for all snippets.
Then, we generate various temporal intervals by collecting the locations of high-score snippets in the score maps.
To obtain temporal action proposals, we estimate the action confidence scores of these temporal intervals and adjust their boundaries.
}
\end{figure}
\IEEEpubidadjcol

Temporal action proposal generation methods \cite{gao2017iccv, shou2016cvpr, zhao2017iccv, buch2017cvpr, escorcia2016eccv, gao2018eccv, lin2018eccv, liu2019cvpr} can be categorized into segment-based, snippet score-based, and hybrid methods.
Segment-based methods \cite{gao2017iccv, shou2016cvpr, buch2017cvpr, escorcia2016eccv} directly use sliding windows to define temporal segments and evaluate their action confidence scores.
By adopting a large number of segments with several temporal spans, these methods achieve high recall performance.
However, uniformly distributed segments provide inaccurate boundaries for action instances.

To address the inaccurate boundary problem of segment-based proposals, snippet score-based methods \cite{lin2018eccv, zhao2017iccv} have been proposed, where a snippet is defined as a set of a few consecutive frames \cite{gao2017iccv}.
In such methods, actionness or boundary scores are evaluated for individual snippets, and the locations of snippets with high scores are then grouped to form proposals.
Although snippet-level evaluation can provide accurate boundaries, low-quality snippet scores often result in missing action instances.

Hybrid methods \cite{liu2019cvpr, gao2018eccv} aim to generate proposals with both high recall and accurate boundaries by combining segment- and snippet score-based methods.
These methods achieve higher performance than other types of methods.
However, hybrid methods still exploit the low-quality 1D score sequences for fusion processes.

To overcome the problems mentioned above, in this paper, we present a novel concept of ``snippet relatedness'' to consider the local relations among snippets.
The snippet relatedness is related to the local consistency of consecutive frames, which plays an important role in tasks requiring the relations among frames or images \cite{wu2018tcsvt,lai2018eccv,dechter1997tcs}.
Specifically, we propose a new snippet score-based approach, named Snippet Relatedness-based Generator (SRG), with the snippet relatedness (see Fig. \ref{fig1}).
Snippet relatedness indicates which snippets are related to a specific action instance.
Based on the snippet relatedness, we evaluate relatedness scores that indicate whether neighboring snippets $\bm{s}_{nbr}=(\bm{s}^{t^-}_{nbr}, \bm{s}^{t^+}_{nbr})$ belong to the same action instance as a reference snippet $s_{ref}$ (see the middle of Fig. \ref{fig1}).
Here, $\bm{s}^{t^-}_{nbr}$ and $\bm{s}^{t^+}_{nbr}$ denote the past and future snippets of $s_{ref}$, respectively.
Note that our relatedness score is different from the actionness score.
First, a relatedness score sequence considers only the action instance to which the reference snippet belongs, while an actionness score sequence includes all action instances in an input video.
Second, existing snippet score-based methods directly use a 1D actionness score sequence to obtain proposals.
Unlike this, we produce a 2D relatedness score map by manipulating the multiple 1D score sequences for all snippets. 
To effectively learn the snippet relatedness, we present ``pyramid non-local (PN) operations'' that locally and globally capture long-range dependencies among snippets.
By employing these components, SRG first produces three types of 2D score maps (i.e., relatedness, starting, and ending), which enable the generation of various snippet score-based temporal intervals.
These temporal intervals reliably cover most action instances with high overlap.
Next, SRG evaluates the action confidence scores of the temporal intervals and adjusts their boundaries to generate temporal action proposals.
In experiments, SRG outperforms state-of-the-art methods for temporal action proposal generation on THUMOS-14 \cite{jiang2015url} and ActivityNet-1.3 \cite{heilbron2015cvpr}.
For temporal action detection, we further demonstrate that with the same a standard action classifier, our proposals outperform competing proposals by a large margin.

In summary, our contributions are four-fold:
\renewcommand\labelitemi{\tiny$\bullet$}
\begin{itemize}
\item We propose a new snippet score-based approach, named SRG, with a novel concept of ``snippet relatedness'' for high-quality temporal action proposal generation.

\item Different from actionness scores, we manipulate multiple 1D score sequences to produce a 2D relatedness score map that enables the generation of various temporal intervals covering most action instances with high overlap.

\item We present ``pyramid non-local (PN) operations'' to effectively learn the snippet relatedness by locally and globally capturing long-range dependencies.

\item We conduct extensive experiments on large-scale datasets, where SRG achieves state-of-the-art performance for temporal action proposal generation.
\end{itemize}

The rest of this paper is organized as follows.
In Section II, we review previous works on action recognition, temporal action detection, and temporal action proposal generation.
In Section III, the details of the proposed method are delineated in two subsections.
Experiments on benchmark datasets are described to verify the effectiveness and superiority of our approach in Section IV. 
In Section V, we discuss the limitations of the proposed method.
Finally, we conclude this paper in Section VI.

\section{Related Works}
\subsection{Action Recognition}
Action recognition on a trimmed video has been studied extensively in the past.
Earlier works \cite{efros2003iccv, jia2008cvpr, wang2011cvpr} exploited hand-crafted features.
Recently, convolutional neural networks (CNNs) have driven impressive progress.
Among the various architectures of CNNs, two-stream networks \cite{carreira2017cvpr, feichtenhofer2016cvpr, ng2015cvpr, simonyan2014nips, wang2016eccv} have achieved state-of-the-art performance by either separately or jointly learning appearance and motion features.
3D CNNs \cite{hara2018cvpr, tran2015iccv} have also achieved excellent performance by learning spatio-temporal features.

\begin{figure*}[t!]
\small
\begin{minipage}[t]{0.785\linewidth}
\centering{\includegraphics[width=.99\linewidth]{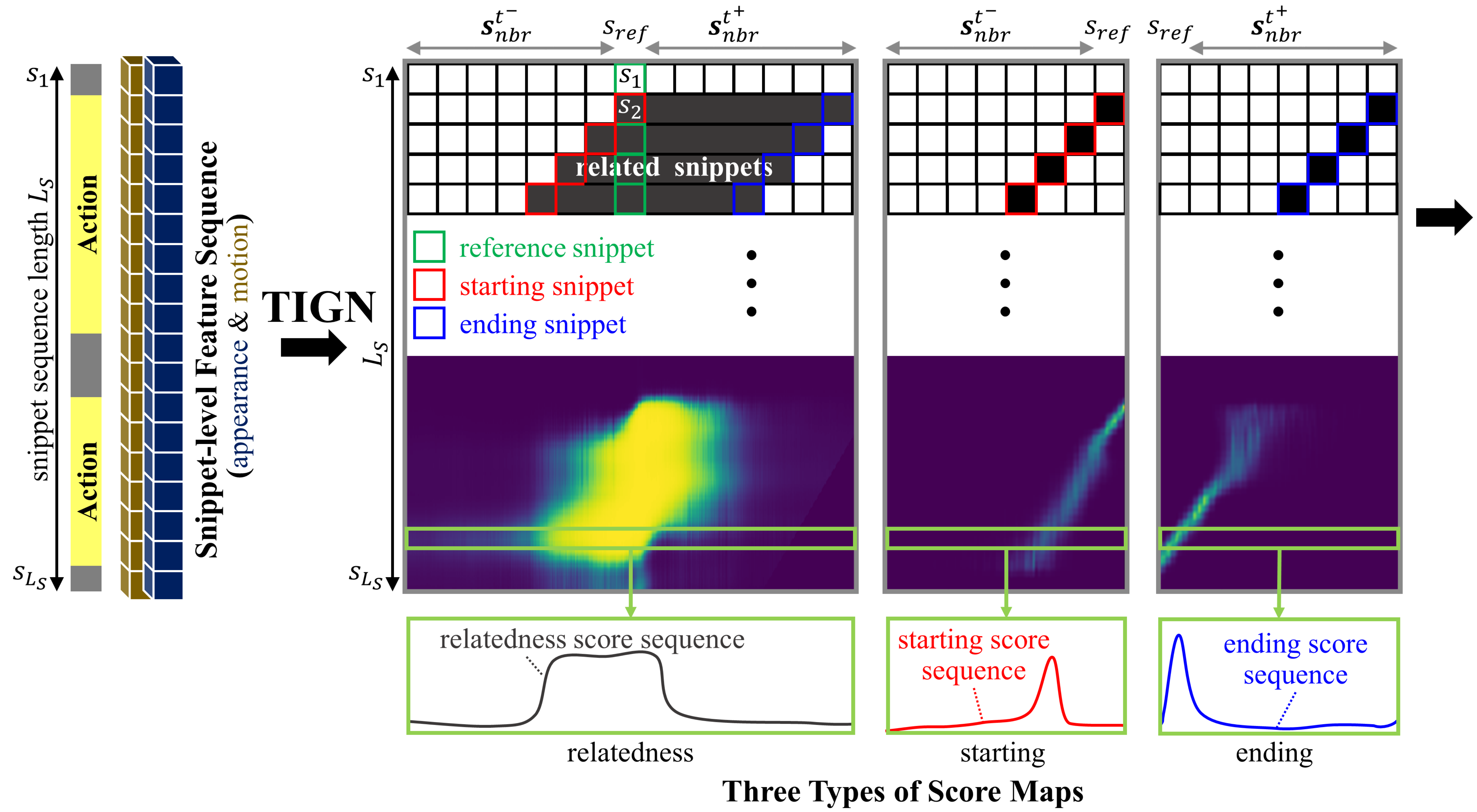}} 
\end{minipage}
\begin{minipage}[t]{0.210\linewidth}
{\centering{\includegraphics[width=.99\linewidth]{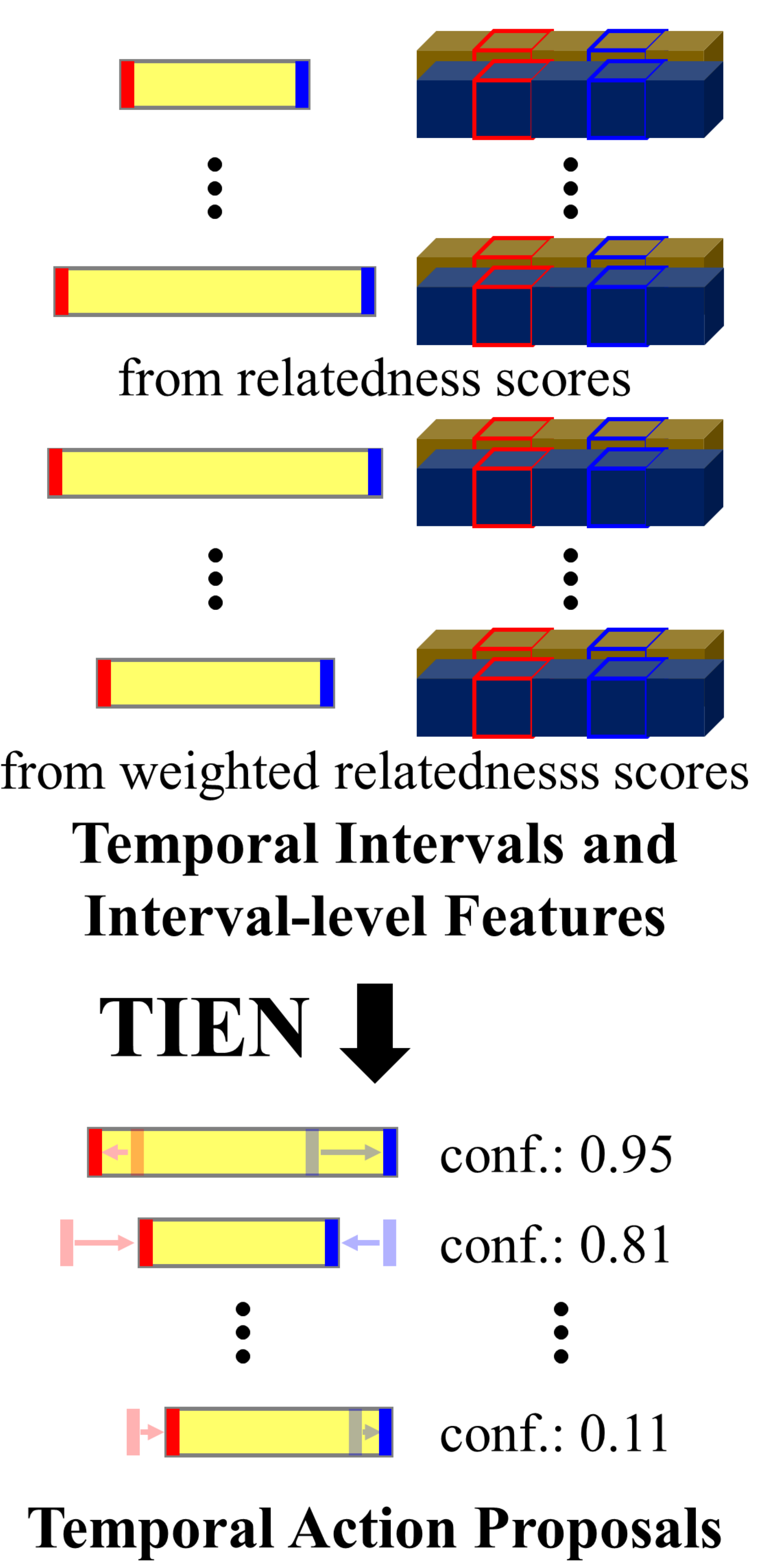}}}
\end{minipage}
\hspace*{4.2cm} (a) Temporal interval generation
\hspace{4.9cm}  (b) Temporal interval evaluation
\caption{\label{fig2}Framework of our approach.
(a) Given a snippet-level feature sequence of an input video, Temporal Interval Generation Network (TIGN) first produces relatedness, starting, and ending score maps.
In each score map, each row describes a score sequence with respect to a reference snippet $s_{ref}$ and its neighboring snippets $\bm{s}_{nbr}=( \bm{s}^{t^-}_{nbr}, \bm{s}^{t^+}_{nbr})$.
Then, snippet score-based temporal intervals are generated based on the relatedness and weighted relatedness score sequences.
(b) By using the temporal intervals and the interval-level features, Temporal Interval Evaluation Network (TIEN) evaluates the action confidence scores of the temporal intervals and refines their boundaries to define temporal action proposals.}
\end{figure*}

\subsection{Temporal Action Detection}
Temporal action detection predicts the action categories, start times, and end times of action instances in an untrimmed video.
S-CNN \cite{shou2016cvpr} uses multiple 3D CNNs consisting of proposal, classification, and localization networks for temporal action detection.
R-C3D \cite{xu2017iccv} and TAL-Net \cite{chao2018cvpr} adopt the framework of Faster R-CNN \cite{ren2015nips} with a modification for considering large variations in action duration.
Shou \emph{et al.} \cite{shou2017cvpr} proposed convolutional-de-convolutional filters that perform spatial downsampling and temporal upsampling operations to learn high-level action semantics and temporal dynamics.
Gao \emph{et al.} \cite{gao2017bmvc} employed cascaded boundary regressors to improve the temporal action boundaries.
Zhao \emph{et al.} \cite{zhao2017iccv} proposed a structured temporal pyramid and a completeness classifier to produce a global action representation.

\subsection{Temporal Action Proposal Generation}
We categorize temporal action proposal generation methods into three types: segment-based, snippet score-based, and hybrid methods. 
Segment-based methods directly use sliding windows for temporal segments and evaluate their action confidence scores.
S-CNN-prop \cite{shou2016cvpr} adopts C3D \cite{tran2015iccv} as a binary classifier to evaluate confidence scores of segments.
TURN \cite{gao2017iccv} jointly conducts binary classification and temporal boundary regression on segments.

Snippet score-based methods are based on snippet-level action scores, which yield accurate action boundaries.
TAG \cite{zhao2017iccv} initially evaluates actionness scores of snippets.
After that, the locations of high-score snippets are grouped as proposals by using the watershed algorithm \cite{roerdink2000fi}.
BSN \cite{lin2018eccv} estimates the boundary scores of snippets to generate proposals, and then their confidence scores are evaluated.

Hybrid methods integrate both types of methods discussed above to generate proposals with both high recall and accurate boundaries.
CTAP \cite{gao2018eccv} first generates initial proposals by employing actionness scores and sliding window sampling.
Then, complementary proposals are collected from the initial proposals.
CTAP further conducts action confidence evaluation and boundary adjustment on the collected proposals.
MGG \cite{liu2019cvpr} consists of a segment proposal generator and a frame actionness generator for segment-based proposals and snippet-level actionness scores, respectively.
To obtain the final proposals, the boundaries of the segment-based proposals are adjusted based on the snippet-level scores.

\section{Our Approach}
In this section, we explain the proposed approach, namely SRG, in two steps: temporal interval generation and temporal interval evaluation.
The framework of our approach is illustrated in Fig. \ref{fig2}.

\subsection{Temporal Interval Generation}
In this step, we aim to generate various snippet score-based temporal intervals with high recall and accurate boundaries.
To this end, we introduce a CNN model, called Temporal Interval Generation Network (TIGN), which produces three types of 2D score maps (i.e., relatedness, starting, and ending) for an input video.

\textbf{Snippet-level feature sequence.}
Similar to previous methods \cite{gao2018eccv, gao2017bmvc}, we define a snippet sequence $S=\{s_l\}^{L_S}_{l=1}$ of an input video consisting of $N_V$ frames, where $L_S$ is the length of $S$.
$L_S$ is calculated as $N_V/N_s$, where $N_s$ is the number of frames in a snippet.
By using a snippet $s_l$ as input of the two-stream network \cite{wang2016eccv}, we obtain an appearance feature vector $f^a_l \in \mathbb{R}^{d_a}$ and a motion feature vector $f^m_l \in \mathbb{R}^{d_m}$, where $d_a$ and $d_m$ are the dimensions of each feature vector. 
Then, a snippet-level feature sequence $F_S$ is denoted as follows:
\begin{eqnarray}
F_S = ( (f^a_1, \cdot\cdot\cdot, f^a_{L_S}), (f^m_1, \cdot\cdot\cdot, f^m_{L_S}) ).
\end{eqnarray}

\begin{figure}[t!]
\small
\begin{minipage}[t]{0.99\linewidth}
\centering{\includegraphics[width=.99\linewidth]{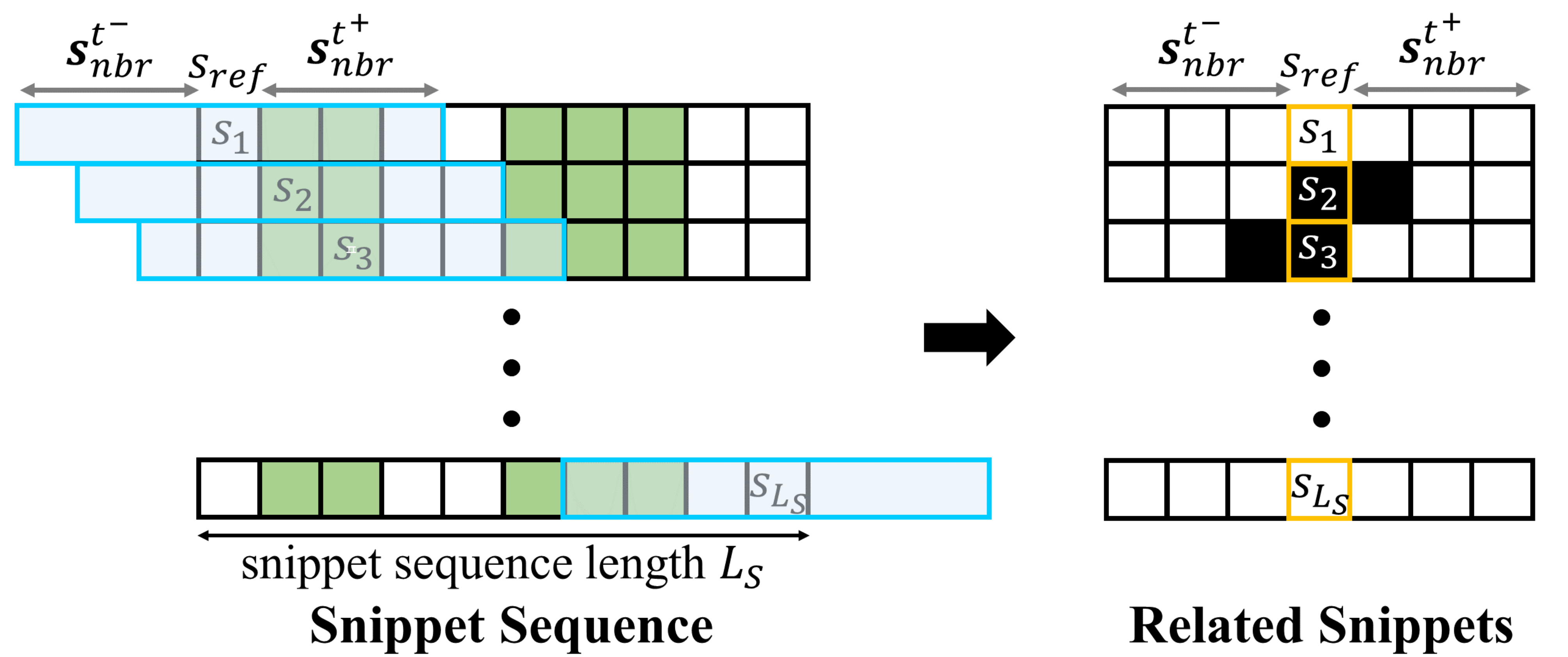}} 
\centering{(a) Related snippets}
\end{minipage}
\begin{minipage}[t]{0.99\linewidth}
{\centering{\includegraphics[width=.99\linewidth]{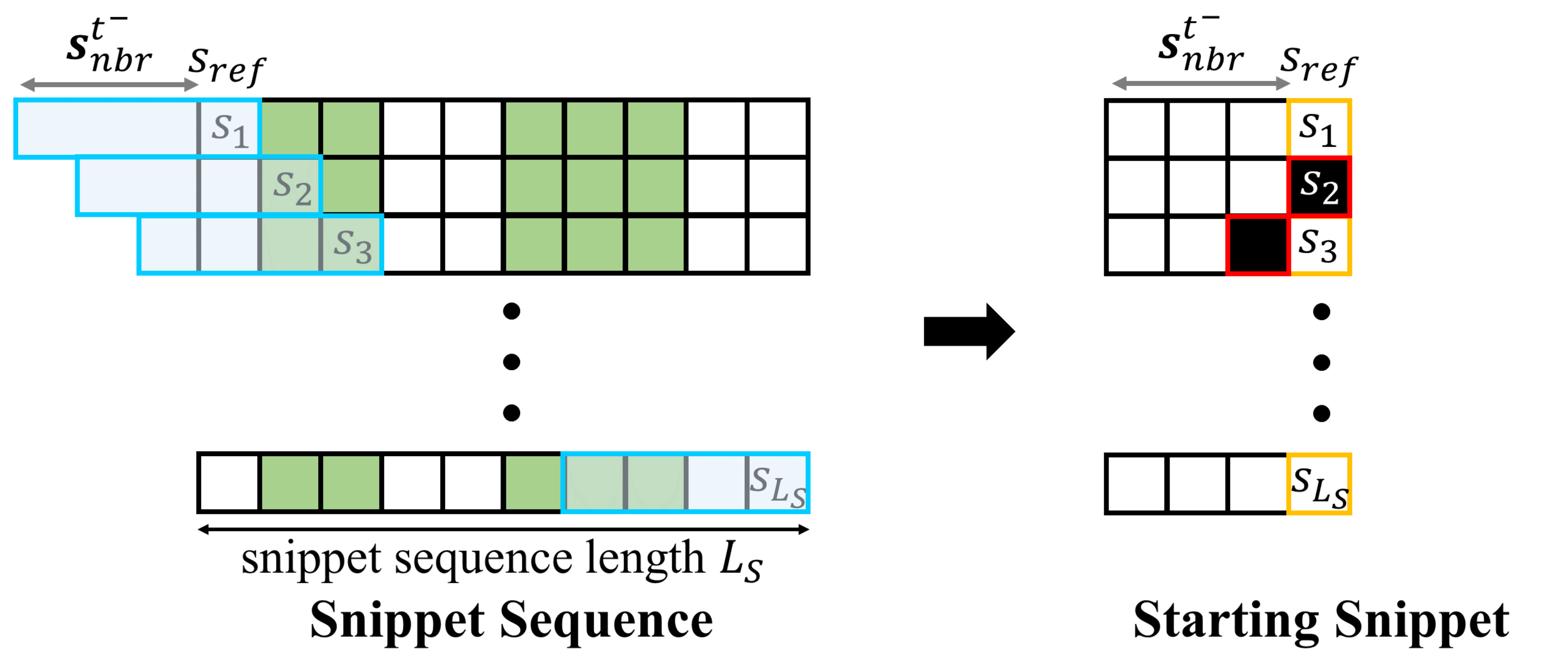}}}
\centering{(b) Starting snippet}
\end{minipage}
\begin{minipage}[t]{0.99\linewidth}
{\centering{\includegraphics[width=.99\linewidth]{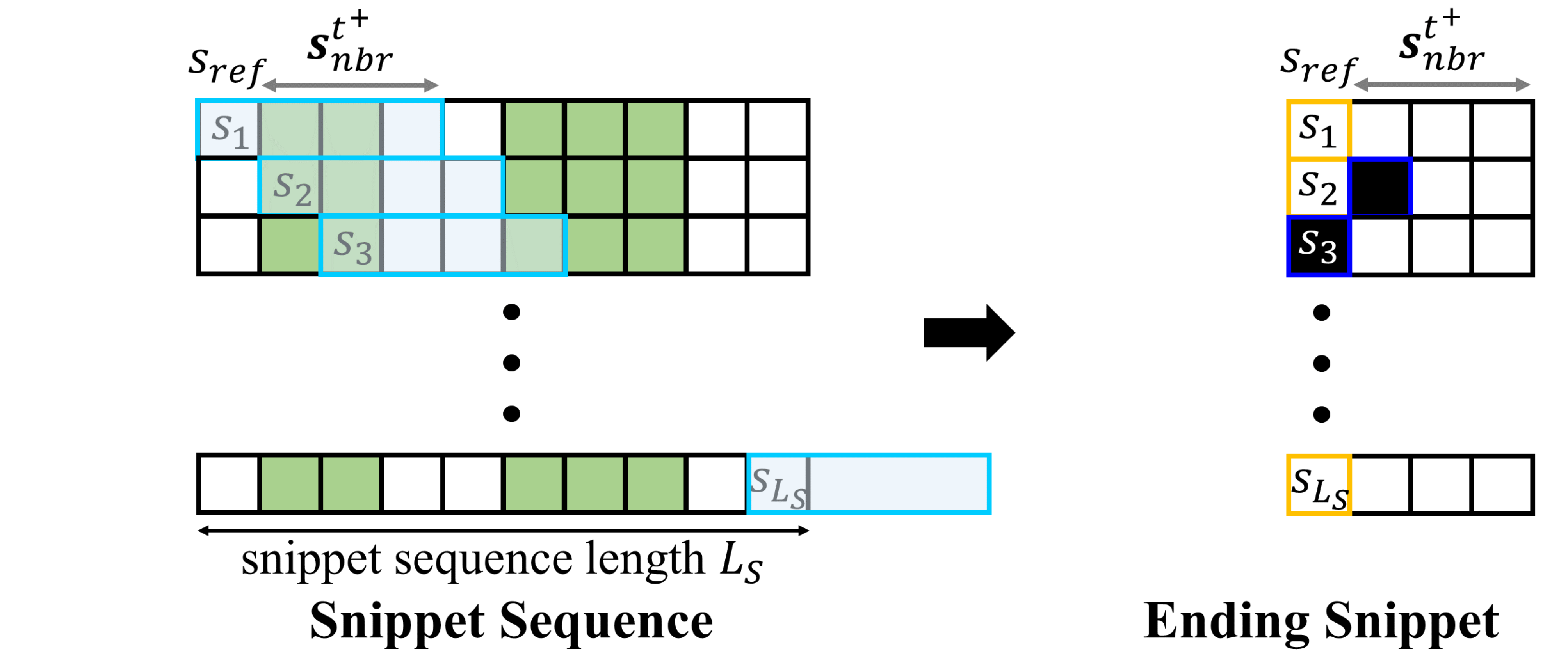}}}
\centering{(c) Ending snippet}
\vspace{0.15cm}
\end{minipage}
\caption{\label{fig3}Toy examples of three types of snippets. We first define a reference snippet $s_{ref}$ and its neighboring snippets $\bm{s}_{nbr} = (\bm{s}^{t^-}_{nbr}, \bm{s}^{t^+}_{nbr})$. Consecutive snippets in green represent an action instance and each region in sky blue represents the currently considered snippets. (a) Related snippets are the snippets that belong to the same action instance as $s_{ref}$. (b) The starting snippet is the starting of the action instance that $s_{ref}$ belongs to. (c) The ending snippet is the ending of the action instance that $s_{ref}$ belongs to.}
\end{figure}

\textbf{Temporal Interval Generation Network (TIGN).}
Given $F_S$ as input, TIGN produces relatedness, starting, and ending score maps.
In detail, the relatedness scores indicate whether neighboring snippets $\bm{s}_{nbr}=(\bm{s}^{t^-}_{nbr}, \bm{s}^{t^+}_{nbr})$ of a reference snippet $s_{ref}$ are related snippets.
We define related snippets as snippets in $\bm{s}_{nbr}$ that belong to the same action instance as $s_{ref}$.
Similar to the relatedness score, the starting (ending) score represents whether the locations of $\bm{s}^{t^-}_{nbr}$ ($\bm{s}^{t^+}_{nbr}$) and $s_{ref}$ are the starting (ending) of the action instance that $s_{ref}$ belongs to.
For clarity, we present toy examples to illustrate the concepts of related, starting, and ending snippets in Fig. \ref{fig3}.
We consider all snippets in $F_S$ as reference snippets, which yields a 1D score sequence for each snippet.
Consequently, TIGN generates three types of 2D score maps by manipulating the corresponding types of multiple 1D score sequences.
Note that TIGN does not use the sliding window technique to generate these 2D score maps, but produces them by processing a snippet sequence for the whole input video at once without any redundant operations.

\begin{figure}[t!]
\centering{\includegraphics[width=.99\linewidth]{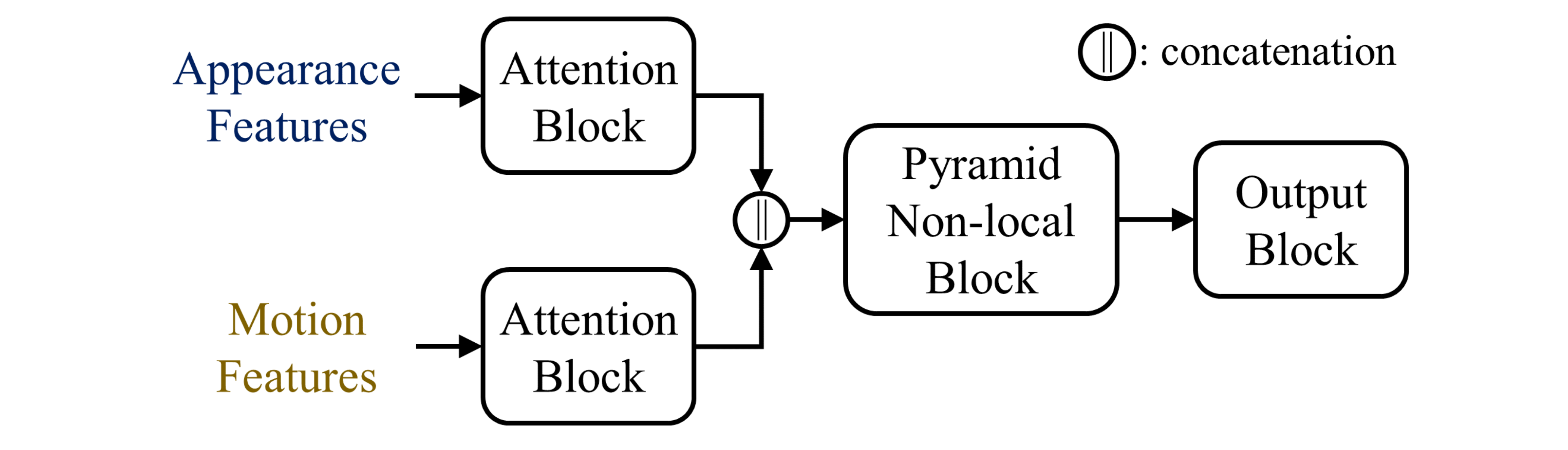}}\\
\caption{\label{fig4}Overview of the architectures of both TIGN and TIEN.}
\end{figure}

\textbf{TIGN architecture.}
As shown in Fig. \ref{fig4}, TIGN consists of three types of blocks, namely, attention, PN, and output blocks.
Motivated by recent successful works on attention \cite{hu2018cvpr, woo2018eccv}, we introduce two attention blocks to focus on important appearance and motion features, individually.
For each attention block, we adopt CBAM \cite{woo2018eccv} and change the spatial attention to temporal attention.

\begin{figure}[t!]
\centering{\includegraphics[width=.99\linewidth]{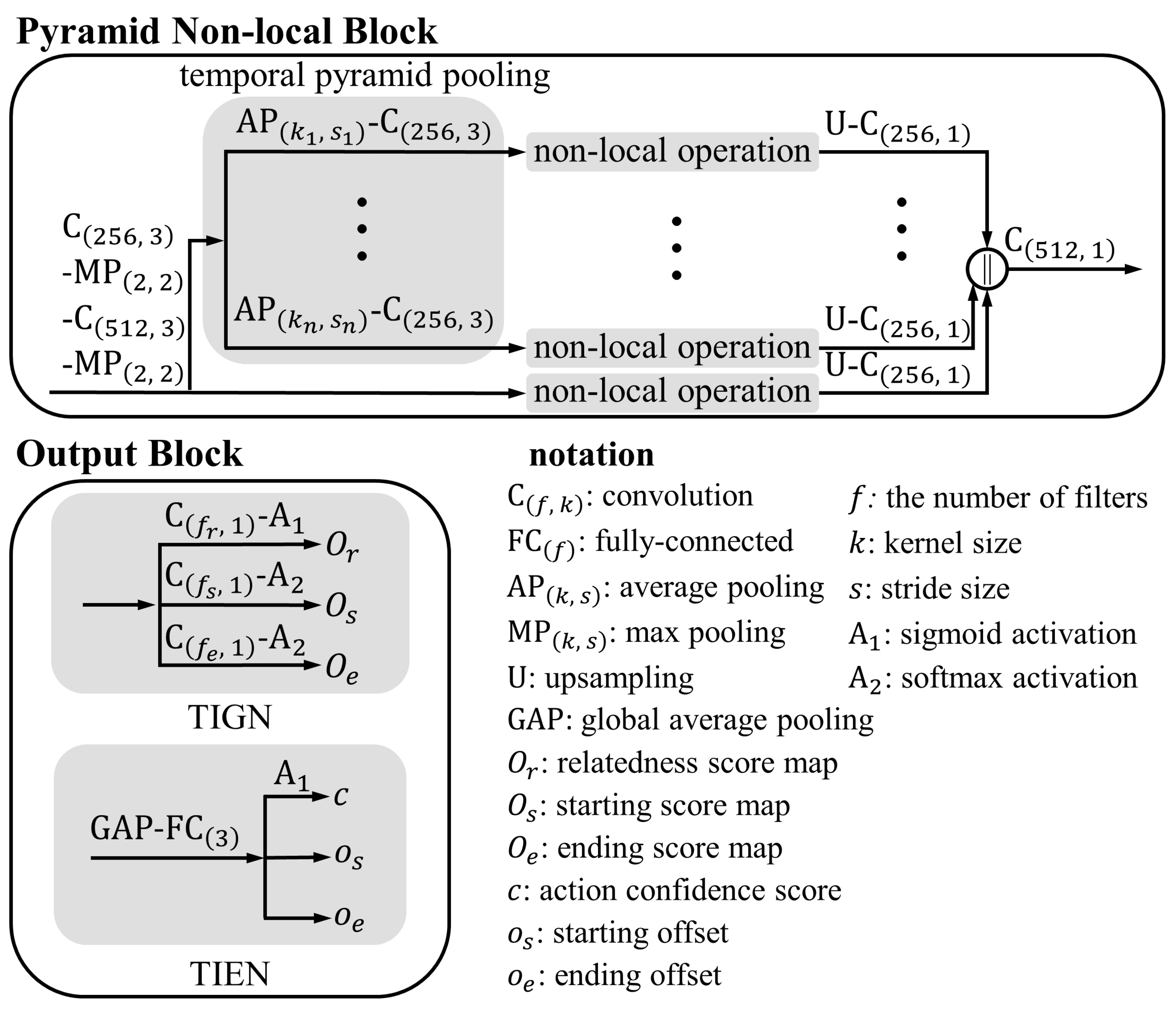}}\\
\caption{\label{fig5}Description of the architectures of pyramid non-local (PN) and output blocks.}
\end{figure}

Given the concatenated outputs of each attention block as input, the PN block performs a temporal pyramid pooling operation and the non-local operation \cite{wang2018cvpr} (see the top of Fig. \ref{fig5}).
In \cite{zhao2017cvpr}, spatial pyramid pooling is successfully used to encode local and global context information.
Inspired by this, we introduce a temporal pyramid pooling operation involving average pooling over several temporal spans.
Furthermore, we apply the non-local operation to each temporal pooling feature and the residual feature to effectively encode the relatedness among all snippets.
Then, we upsample the outputs of the non-local operations to the length of the input snippet sequence.
Finally, we concatenate these upsampled features to be fed into a convolutional layer.

In the output block, we employ three convolutional layers and three activation functions to estimate a relatedness score map $O_r$, a starting score map $O_s$, and an ending score map $O_e$ (see the bottom left of Fig. \ref{fig5}).
For the three convolutional layers, the numbers of filters $f_r$, $f_s$, and $f_e$ are set to the lengths of $O_r$, $O_s$, and $O_e$, respectively.
We use sigmoid activation for $O_r$ and softmax activation for $O_s$ and $O_e$.

\begin{figure}[t!]
\centering{\includegraphics[width=.99\linewidth]{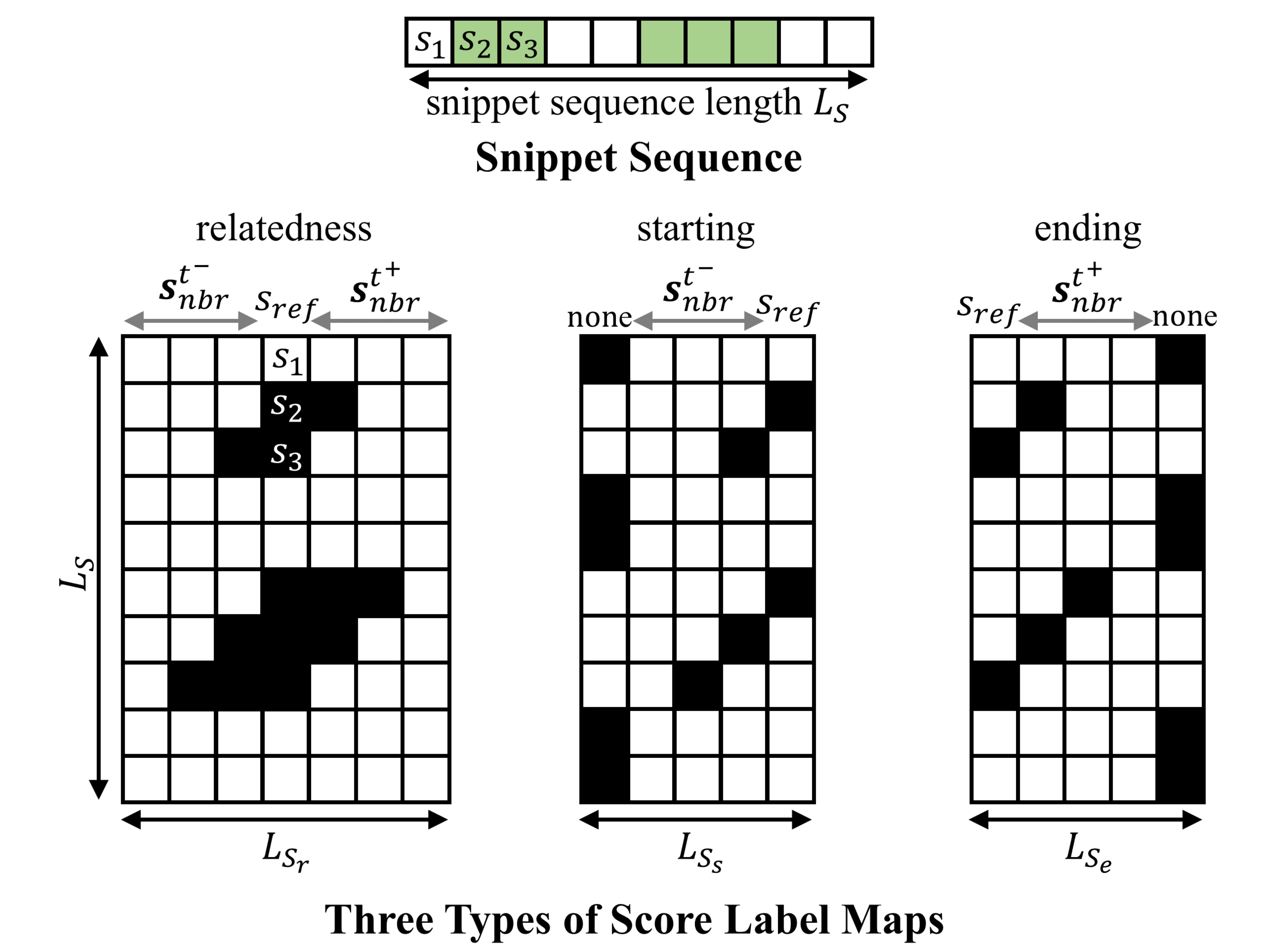}}\\
\caption{\label{fig6}Toy examples of the annotation of relatedness, starting, and ending score label maps. The index ``none'' indicates that an action instance including $s_{ref}$ does not exist.}
\end{figure}

\textbf{TIGN training.}
For training TIGN, we design a multi-task loss function $L^{TIGN}$ as follows:
\begin{eqnarray}
L^{TIGN} = L^{TIGN}_r + L^{TIGN}_s + L^{TIGN}_e,
\end{eqnarray}
where $L^{TIGN}_r$, $L^{TIGN}_s$, and $L^{TIGN}_e$ are loss functions defined for the relatedness, starting, and ending score maps, respectively.
We adopt a binary cross-entropy loss for $L^{TIGN}_r$ and a multi-class cross-entropy loss for $L^{TIGN}_s$ and $L^{TIGN}_e$ as follows:
\begin{eqnarray}
\begin{aligned}
L^{TIGN}_r = &- \frac{1}{L_S L_{S_r}} \sum^{L_S}_{i=1} \sum^{L_{S_r}}_{j=1} \Big( M_{r,ij}log(O_{r,ij})  \\
&  + (1 - M_{r,ij}) log(1 - O_{r,ij}) \Big), \\
\end{aligned} \\
L^{TIGN}_s = - \frac{1}{L_S} \sum^{L_S}_{i=1} \sum^{L_{S_s}}_{j=1} \Big( M_{s,ij}log(O_{s,ij}) \Big), \\
L^{TIGN}_e = - \frac{1}{L_S} \sum^{L_S}_{i=1} \sum^{L_{S_e}}_{j=1} \Big( M_{e,ij}log(O_{e,ij}) \Big),
\end{eqnarray}
where $M_r \in \mathbb{R}^{L_S \times L_{S_r}}$, $M_s \in \mathbb{R}^{L_S \times L_{S_s}}$, and $M_e \in \mathbb{R}^{L_S \times L_{S_e}}$ are label maps for the relatedness, starting, and ending scores, respectively, and $L_{S_r}$, $L_{S_s}$, and $L_{S_e}$ are the widths of the corresponding label maps.
We annotate $M_r$, $M_s$, and $M_e$ based on action instances in a snippet sequence, as shown in Fig. \ref{fig6}.
$M_r$ can be annotated as previously described with regard to the concept of the snippet relatedness.
For $M_s$ and $M_e$, we add the index ``none'' for the non-existence of an action instance to enable the use of softmax activation for each row.

\begin{figure}[t!]
\centering{\includegraphics[width=.99\linewidth]{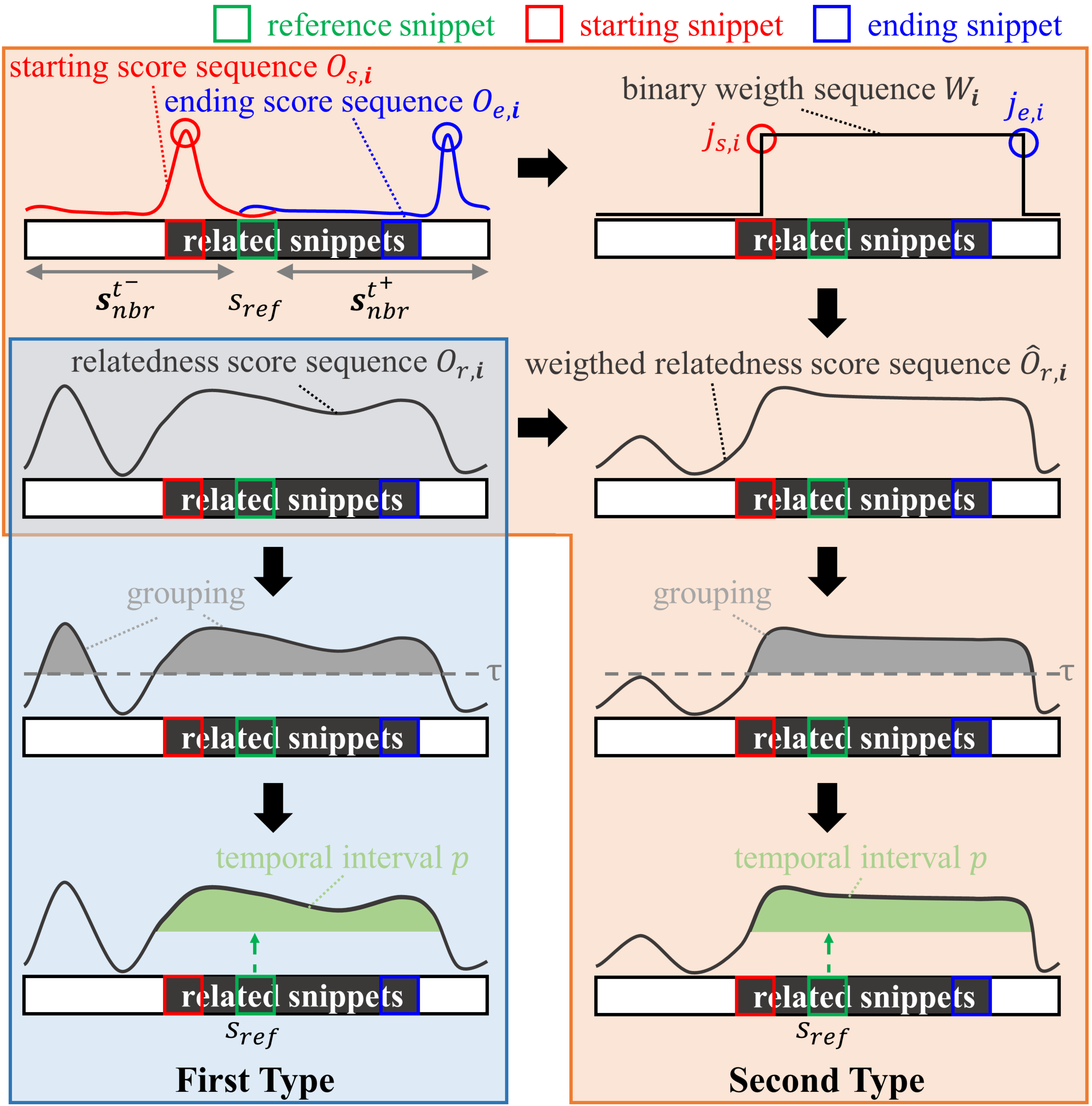}}\\
\caption{\label{fig7}Two types of temporal interval generation. We generate temporal intervals based on a relatedness score sequence and a weighted relatedness score sequence.}
\end{figure}

\textbf{Temporal intervals.}
We generate two types of temporal intervals $p=(t_{s}, t_{e})$ based on the score sequences in $O_r$, $O_s$, and $O_e$, as shown in Fig. \ref{fig7}.
Here, $t_{s}$ and $t_{e}$ are the starting and ending locations, respectively, of the temporal interval.
Let $O_{r,\bm{i}}$, $O_{s,\bm{i}}$, and $O_{e,\bm{i}}$ denote the $i$th score sequences in $O_r$, $O_s$, and $O_e$, respectively.

Given $O_{r,\bm{i}}$, for the first type of temporal interval, we initially obtain the temporal locations of high-score snippets by performing a thresholding operation with a threshold value of $\tau$.
We then group the consecutive temporal locations of high-score snippets.
Among the location groups, we select only the location group that includes $s_{ref,i}$ to collect $p_i=(t_{s,i}, t_{e,i})$.
We apply this process to all relatedness score sequences in $O_r$ to obtain as many reliable temporal intervals as possible, as sketched in Alg. 1.

\begin{algorithm}[t!]
\caption{Temporal Interval Generation 1}
\label{alg1}
\SetKwInOut{Input}{Input}
\SetKwInOut{Initialization}{Initialization}
\SetKwInOut{Output}{Output}
\Input{a relatedness score map $O_r$, a threshold value $\tau$}
\For{$i = 1$ \KwTo $L_S$}{
\For{$j = 1$ \KwTo $L_{S_r}$}{
\If{$O_{r,ij} \geq \tau$}{add $j$ to $\bm{j}$}
}
group consecutive temporal locations in $\bm{j}$ \\
select the location group including $s_{ref,i}$ \\
collect $p_i=(t_{s,i}, t_{e,i})$ of the selected group
}
\Output{a set of temporal intervals $P$}
\end{algorithm}

\begin{algorithm}[t!]
\caption{Temporal Interval Generation 2}
\label{alg1}
\SetKwInOut{Input}{Input}
\SetKwInOut{Initialization}{Initialization}
\SetKwInOut{Output}{Output}
\Input{a relatedness score map $O_r$, a starting score map $O_s$, a ending score map $O_e$, a threshold value $\tau$}
\Initialization{a weighted relatedness score map $\hat{O}_r \! \leftarrow \! \bm{0}$ \!\!\!\!\!\!}
compute a binary weight map $W$ from $O_s$ and $O_e$\\
\For{$i = 1$ \KwTo $L_S$}{
\For{$j = 1$ \KwTo $L_{S_r}$}{
$\hat{O}_{r,ij} = (O_{r,ij} + W_{ij}) / 2$ \\
\If{$\hat{O}_{r,ij} \geq \tau$}{add $j$ to $\bm{j}$}
}
group consecutive temporal locations in $\bm{j}$ \\
select the location group including $s_{ref,i}$ \\
collect $p_i=(t_{s,i}, t_{e,i})$ of the selected group
}
\Output{a set of temporal intervals $P$}
\end{algorithm}

For the second type of temporal interval, we first obtain the temporal locations of starting and ending snippets with the highest scores in $O_{s,\bm{i}}$ and $O_{e,\bm{i}}$, respectively.
We then generate a binary weight sequence $W_{\bm{i}}$ with values of 0 for locations outside the boundaries and 1 for locations inside the boundaries:
\begin{eqnarray}
W_{ij}=\left\{\begin{matrix}
1, &  \text{if $j_{s,i} \leq j \leq j_{e,i}$} \\
0, &  \text{otherwise}
\end{matrix}\right. ,
\end{eqnarray}
where $j_{s,i}$ and $j_{e,i}$ are the starting and ending locations of the $i$th corresponding score sequences, respectively.
Afterward, we compute a weighted relatedness score sequence $\hat{O}_{r,\bm{i}}$ by combining the relatedness score sequence $O_{r,\bm{i}}$ with the boundary-based binary weight sequence $W_{\bm{i}}$ as follows:
\begin{eqnarray}
\hat{O}_{r,\bm{i}} = (O_{r,\bm{i}} + W_{\bm{i}})/2,
\end{eqnarray}
where the addition and division are element-wise operations.
From $\hat{O}_{r,\bm{i}}$, we then collect a temporal interval $p_i=(t_{s,i},t_{e,i})$ by applying the same rule used for the first type of temporal interval generation (see Alg. 2).

By employing both types of temporal intervals, we define a set of temporal intervals $P=\{p_n\}^{N_P}_{n=1}$, where $N_P$ is the number of temporal intervals.

\subsection{Temporal Interval Evaluation}
In this step, we evaluate the action confidence scores for temporal intervals and adjust their boundaries to generate temporal action proposals.
To that end, we introduce a CNN model, named Temporal Interval Evaluation Network (TIEN), which uses interval-level features for temporal intervals as input.

\textbf{Interval-level features.}
We define an interval-level feature $F_I$ by aggregating snippet-level features within the starting and ending locations of a temporal interval.
Similar to \cite{gao2018eccv, lin2018eccv}, we exploit contextual information by considering boundary regions.
Specifically, we use snippet-level features within [$(t_s - L_C)$, $(t_e + L_C)$] to form $F_I$ as follows:
\begin{eqnarray}
\begin{aligned}
F_I = ((&f^a_{( t_s-L_C )},  \cdot\cdot\cdot, f^a_{t_s}, \cdot\cdot\cdot, f^a_{t_e}, \cdot\cdot\cdot, f^a_{( t_e+L_C )}), \\
&(f^m_{( t_s-L_C )}, \cdot\cdot\cdot, f^m_{t_s}, \cdot\cdot\cdot, f^m_{t_e}, \cdot\cdot\cdot, f^m_{( t_e+L_C )})),
\end{aligned}
\end{eqnarray}
where $L_C$ is the length of contextual snippets in the boundary region.

\textbf{Temporal Interval Evaluation Network (TIEN).}
Given a set of interval-level features as input, TIEN performs action confidence evaluation and boundary adjustment.
Interval-level evaluation provides more reliable results than snippet-level evaluation due to the consideration of global information.
As the output of TIEN, we obtain the action confidence score, starting offset, and ending offset for each temporal interval.

\textbf{TIEN architecture.}
We construct TIEN using attention, PN, and output blocks (see Fig. \ref{fig4}).
The attention and PN blocks are the same as those of TIGN.
For the output block, we use a global average pooling layer, a fully connected layer, and sigmoid activation to obtain an action confidence score $c$, a starting offset $o_s$, and an ending offset $o_e$ for each temporal interval (see the bottom left of Fig. \ref{fig5}).

\textbf{TIEN training.}
To train TIEN, we use temporal intervals generated by TIGN as training samples.
We calculate temporal intersection-over-union (tIoU) scores between the temporal intervals and action instances to define action confidence scores $c^g$ as the ground truth.
Among all of the temporal intervals, we use those with $c^g \geq 0.5$ as positive samples and those with $c^g \leq 0.1$ as negative samples.
The loss function $L^{TIEN}$ is defined as follows:
\begin{eqnarray}
L^{TIEN} = L^{TIEN}_c + \alpha (L^{TIEN}_s + L^{TIEN}_e),
\end{eqnarray}
where $L^{TIEN}_c$, $L^{TIEN}_s$, and $L^{TIEN}_e$ are loss functions defined for the action confidence scores, starting offsets, and ending offsets, respectively.
$\alpha$ is a parameter for balancing the contributions among the loss functions, which is empirically set to 0.1.
We adopt the $L_1$ distance for all three loss functions as follows:
\begin{eqnarray}
L^{TIEN}_c = \frac{1}{N_P} \sum^{N_P}_{n=1} ( |c_n - c^g_n | ),  \\
L^{TIEN}_s = \frac{1}{N_P} \sum^{N_P}_{n=1} c^{g'}_n ( |o_{s,n} - o^g_{s,n} | ), \\
L^{TIEN}_e = \frac{1}{N_P} \sum^{N_P}_{n=1} c^{g'}_n ( |o_{e,n} - o^g_{e,n} | ),
\end{eqnarray}
where $c^{g'}_n$ is a binary label that is set to 1 for a positive sample and  0 for a negative sample.

\textbf{Temporal action proposals.}
By using the output of TIEN, we newly generate a set of temporal action proposals denoted by $P=\{p_n\}^{N_P}_{n=1}$, where $p_n=(t_{s,n}, t_{e,n}, o_{s,n}, o_{e,n}, c_n)$.
To remove redundant proposals, we additionally perform non-maximum suppression (NMS) in terms of temporal overlap and action confidence scores.

\section{Experiments}
\subsection{Datasets}
\textbf{THUMOS-14.} THUMOS-14 \cite{jiang2015url} contains 1,010 and 1,574 untrimmed videos in validation and test sets, respectively.
Among them, 200 and 212 videos in the validation and test sets, respectively, have temporal annotations with 20 action classes.
We trained our models on the validation set and conducted an evaluation on the test set, as done in previous works \cite{gao2017iccv, chao2018cvpr, gao2018eccv, lin2018eccv, liu2019cvpr}.

\textbf{ActivityNet-1.3.} ActivityNet-1.3 \cite{heilbron2015cvpr} consists of 19,994 untrimmed videos containing 200 action classes, where the training, validation, and test sets contain 10,024, 4,926, and 5,044 videos, respectively.
Since annotation for the test set is not publicly available, we trained SRG on the training set and conducted an evaluation on the validation set.

\subsection{Temporal Action Proposal Generation}
\textbf{Snippet-level features.}
We used the spatial and temporal networks in \cite{wang2016eccv} to extract snippet-level features.
Both networks are based on Inception-v3 \cite{szegedy2016cvpr} and were trained on Kinetics-400 \cite{carreira2017cvpr}.
We set the number of frames per snippet to 6 and 18 for THUMOS-14 and ActivityNet-1.3, respectively.

\textbf{Implementation details.}
For THUMOS-14, we set the numbers of both $\bm{s}^{t^-}_{nbr}$ and $\bm{s}^{t^+}_{nbr}$ in TIGN to 600 based on the maximum action length in the training videos.
Thus, $f_r$, $f_s$, and $f_e$ in the output block were set to 1201, 602, and 602, respectively.
For the PN block of TIGN, we empirically set the number of the pyramid level to 4 with kernel and stride size pairs of (3, 1), (5, 3), (7, 5), and (15, 7).
We used the same number of the pyramid level with kernel and stride size pairs of (1, 3), (3, 3), (5, 3), and (7, 3) for the PN block of TIEN.
We trained TIGN for 15 epochs with a learning rate that was initially set to $1\times10^{-4}$ and exponentially decayed every 10 steps with a base value of 0.96.
We trained TIEN for $1\times10^{4}$ steps with the same learning rate strategy used for TIGN.
We used batch sizes of 1 and 256 for TIGN and TIEN, respectively.
For the interval-level features, we set $L_C$ to 20 and rescaled the temporal length of the features to 128.
$L_C$ was adopted to consider contextual information, as in previous works \cite{gao2018eccv,lin2018eccv}.
To select the optimal value of $L_C$, we tested values of $L_C$ such as 0, 5, 10 and 20.
Among these values, $L_C=0$ yielded the worst performance with a relatively high loss value, whereas $L_C=20$ resulted in convergence at a lower loss value compared with $L_C=5$ and $L_C=10$.

\begin{figure}[t!]
\centering
\begin{minipage}[t]{0.76\linewidth}
\includegraphics[width=1.\linewidth]{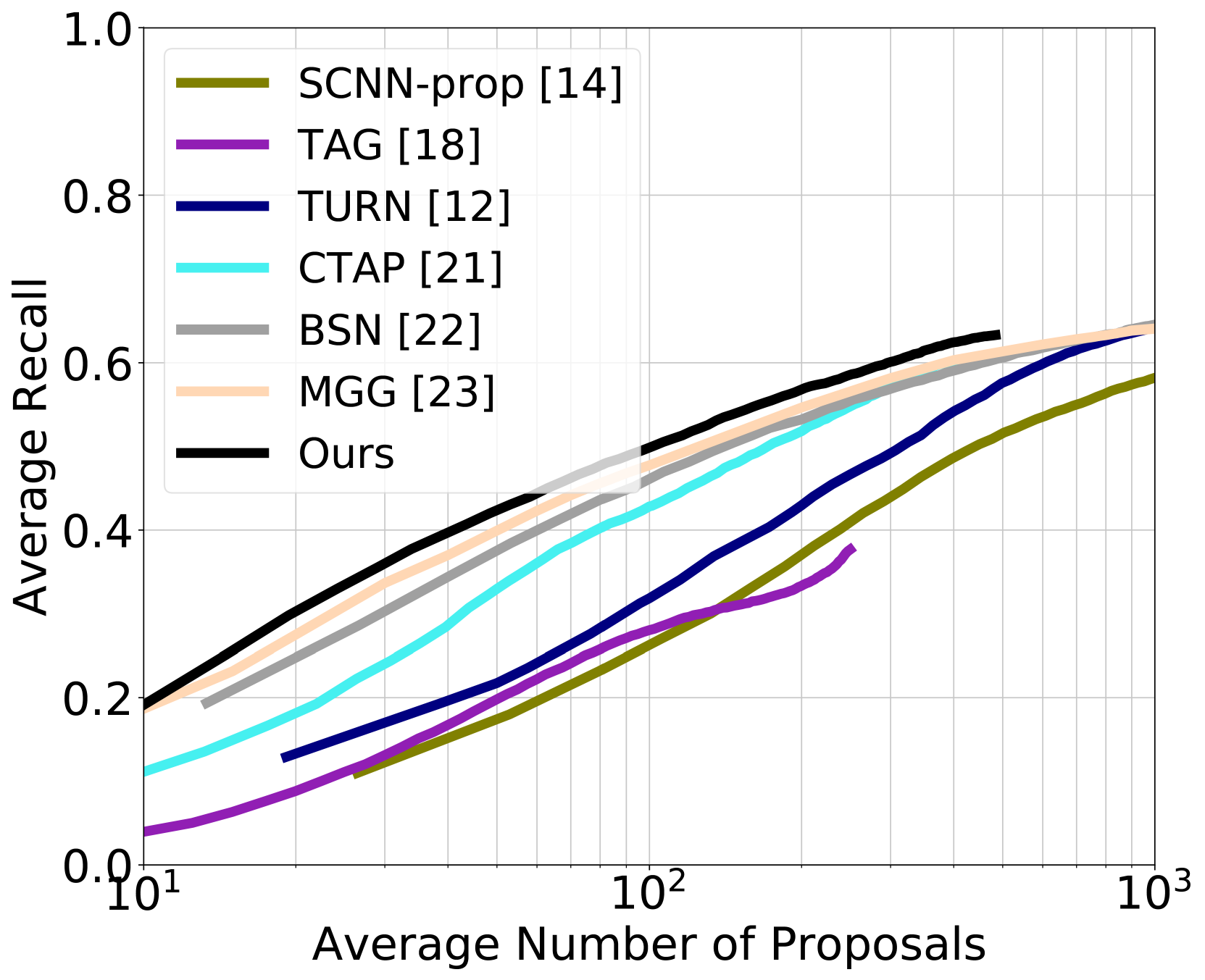}
\end{minipage}
\begin{minipage}[t]{0.73\linewidth}
\includegraphics[width=1.\linewidth]{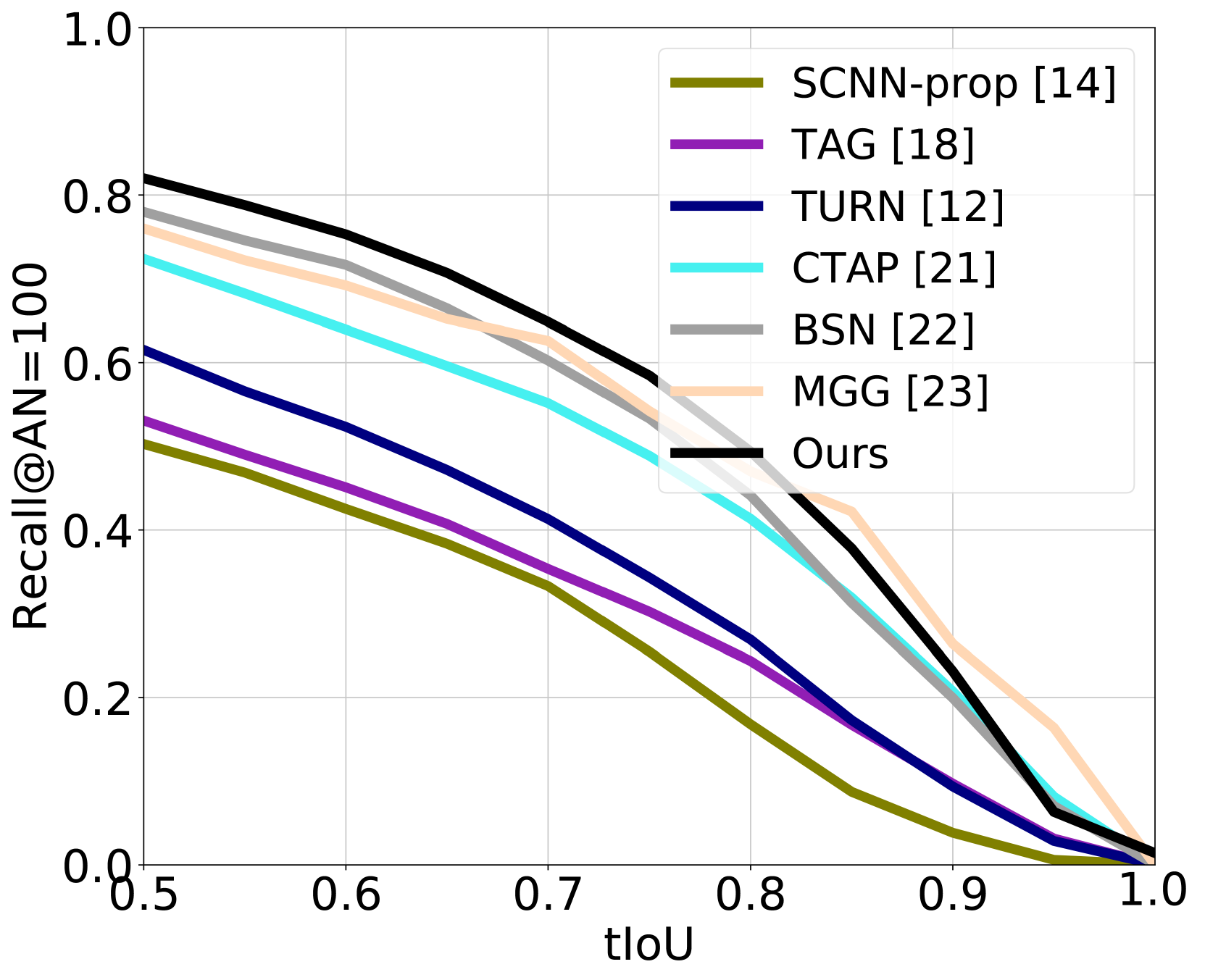}\\
\end{minipage}
\caption{\label{fig8}Performance comparison based on the curves of the average recall (AR) vs. the average number of proposals (AN) (top) and the Recall@AN=100 vs. the temporal intersection-over-union (tIoU) (bottom) on THUMOS-14.}
\end{figure}

For ActivityNet-1.3, we set the numbers of both $\bm{s}^{t^-}_{nbr}$ and $\bm{s}^{t^+}_{nbr}$ in TIGN to 540 to consider the maximum action length in the training set.
Thus, $f_r$, $f_s$, and $f_e$ in the output block were set to 1,081, 542, and 542, respectively.
We used the same kernel and stride sizes used for the PN block on the THUMOS-14 dataset.
The batch sizes for training TIGN and TIEN were also the same as the values used for THUMOS-14.
We trained TIGN for 10 epochs with a learning rate that was initially set to $1\times10^{-4}$ and exponentially decayed every 100 steps with a base value of 0.96.
We trained TIEN for $3\times10^{5}$ steps with the same learning rate strategy used for TIGN.
We rescaled the temporal length of the interval-level features to 128.
We set $L_C$ to 10 because this value empirically yielded the best training results.

For both datasets, we set $\tau$ to 0.1 with an increasing step size of 0.1 to generate temporal intervals based on the relatedness and weighted relatedness score sequences.
Note that, unlike TAG \cite{zhao2017iccv}, CTAP \cite{gao2018eccv}, and MGG \cite{liu2019cvpr} using actionness scores, we do not group multiple temporal intervals since our score sequences consider only the action instance related to the current reference snippet.
For NMS, we empirically set the threshold to 0.83 on THUMOS-14 and adaptively set the threshold to $1 - N_P\times10^{-4}$ on ActivityNet-1.3.

\textbf{Evaluation metrics.}
As done in previous studies, we computed the average recall (AR) with multiple tIoU thresholds. 
We used tIoU thresholds of $[0.50 : 0.05 : 1.00]$ and $[0.50 : 0.05 : 0.95]$ for THUMOS-14 and ActivityNet-1.3, respectively.
On the THUMOS-14 dataset, we measured the AR by varying the average number of proposals (AN) from 50 to 500.
On ActivityNet-1.3, we computed AR@AN=100 and the area under AR vs. AN curve (AUC).
For the AUC, we considered AN values of $[1 : 1 : 100]$.

\textbf{Performance comparison.}
For the THUMOS-14 dataset, we plot the AR vs. AN and Recall@AN=100 vs. tIoU curves of SRG and 6 previous methods, S-CNN-prop \cite{shou2016cvpr}, TAG \cite{zhao2017iccv}, TURN \cite{gao2017iccv}, CTAP \cite{gao2018eccv}, BSN \cite{lin2018eccv}, and MGG \cite{liu2019cvpr}, in Fig. \ref{fig8}, where SRG outperformed state-of-the-art proposal generation methods on both AR vs. AN and Recall@AN=100 vs. tIoU curves.
We further compared SRG to 9 previous methods, DAPs \cite{escorcia2016eccv}, S-CNN-prop , SST \cite{buch2017cvpr}, TURN, TAL-Net-prop \cite{chao2018cvpr}, TAG, CTAP, BSN, and MGG, by evaluating the AR with varying AN.
As shown in Table \ref{tab1}, SRG achieved the best performance among these state-of-the-art methods at every AN.

\begin{table}[t!]
\centering
\caption{\label{tab1}Performance comparison in terms of AR@AN on THUMOS-14.}
\begin{tabular}{C{2.2cm}|C{1.4cm}|C{.55cm} C{.55cm} C{.55cm} C{.55cm}}\hline 
Method & Feature & @50 & @100 & @200 & @500 \\ \hline\hline
DAPs \cite{escorcia2016eccv} & C3D 		& 13.56 	 & 23.83   & 33.96   & 49.29 		\\
S-CNN-prop \cite{shou2016cvpr} & C3D		& 17.22 	 & 26.17   & 37.01   & 51.57 		\\
SST \cite{buch2017cvpr} & C3D			& 19.90   & 28.36   & 37.90   & 51.58 		\\
TURN \cite{gao2017iccv} & Flow 		& 21.86	  & 31.89 & 43.02 & 57.63 	\\
TAL-Net-prop \cite{chao2018cvpr} & Flow	& 35.80	  & 42.30 & 47.50 & - 		\\ 
TAG \cite{zhao2017iccv} & Two-Stream	& 18.55	  & 29.00 & 39.61 & -  	\\
CTAP \cite{gao2018eccv} & Two-Stream 	& 31.03	  & 40.23 & 50.13 & - 		\\
BSN \cite{lin2018eccv} & Two-Stream 	& 37.46	  & 46.06 & 53.21 & 60.64  	\\
MGG \cite{liu2019cvpr} & Two-Stream 	& 39.93    & 47.75 & 54.65 & 61.36  	\\
SRG & Two-Stream 	& \bf{42.19} & \bf{49.72} & \bf{56.71} & \bf{63.78}   \\ \hline
\end{tabular}
\end{table}

\begin{table}[t!]
\centering
\caption{\label{tab2}Performance comparison in terms of AR@AN=100 and the AUC on ActivityNet-1.3.}
\begin{tabular}{C{1.6cm}| C{1.4cm} C{.7cm} C{.7cm} C{.7cm} C{.7cm}}\hline 
Method & SSAD-prop \cite{lin2017acmmm} & CTAP \cite{gao2018eccv} & BSN \cite{lin2018eccv} & MGG \cite{liu2019cvpr} & SRG \\ \hline\hline
AR@AN=100 & 73.01 & 73.17 & 74.16 & 74.56 & \bf{74.65} \\
AUC & 64.40 & 65.72 & 66.17 & \bf{66.54} & 66.06 \\ \hline
\end{tabular}
\end{table}

\begin{table}[t!]
\centering
\caption{\label{tab3a}Comparison of execution times in FPS.}
\begin{tabular}{C{2.8cm}| C{2.1cm}| C{1.1cm}} \hline
Method & Feature & FPS \\ \hline\hline
DAPs \cite{escorcia2016eccv} & C3D & 134 \\
S-CNN-prop \cite{shou2016cvpr} & C3D & 60 \\
SST \cite{buch2017cvpr} & C3D & 301 \\
CTAP \cite{gao2018eccv} & Two-Stream & 418 \\
SRG & Two-Stream & \bf{473} \\ \hline
\end{tabular}
\end{table}

\begin{figure*}[t]
\centering{\includegraphics[width=0.99\linewidth]{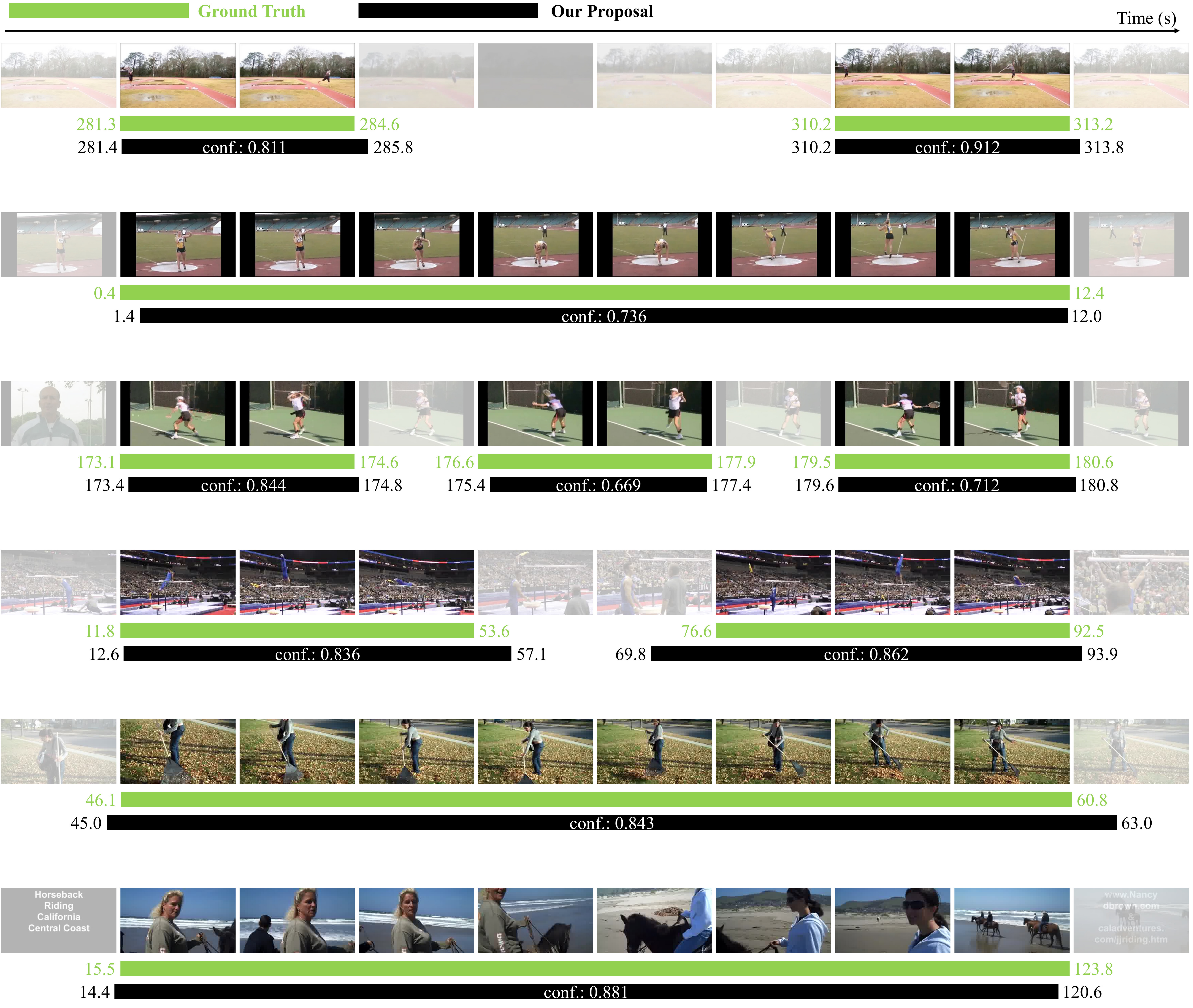}}
\caption{\label{fig9}
Qualitative evaluation of our temporal action proposals on THUMOS-14 (first, second, and third rows) and ActivityNet-1.3 (fourth, fifth, and sixth rows). Clear images represent action frames whereas the images covered by semiopaque masks represent background frames. The left and right numbers of a bar indicate the start and end times, respectively.}
\end{figure*}

On ActivityNet-1.3, we compared SRG to four state-of-the-art proposal generators, SSAD-prop \cite{lin2017acmmm}, CTAP, BSN, and MGG, in terms of AR@AN=100 and the AUC.
As shown in Table \ref{tab2}, SRG outperformed the state-of-the-art methods in terms of AR@AN=100 and achieved comparable performance in terms of the AUC.

For an efficiency analysis, we compared the execution times of the proposed method and four existing approaches, DAPs, S-CNN-prop, SST, and CTAP, in Table \ref{tab3a}.
We measured the processing speed in terms of frames per second (FPS) of the entire process, from feature extraction to temporal interval evaluation, using a Titan X GPU.
We found that SRG operates at 473 FPS, faster than any of the other methods.

\textbf{Qualitative evaluation.}
In Fig. \ref{fig9}, we present a qualitative evaluation of our temporal action proposals on THUMOS-14 and ActivityNet-1.3.
Our proposals reliably capture action instances with accurate boundaries and high action confidence scores.

\begin{figure*}[t!]
\begin{center}
\includegraphics[width=0.83\linewidth]{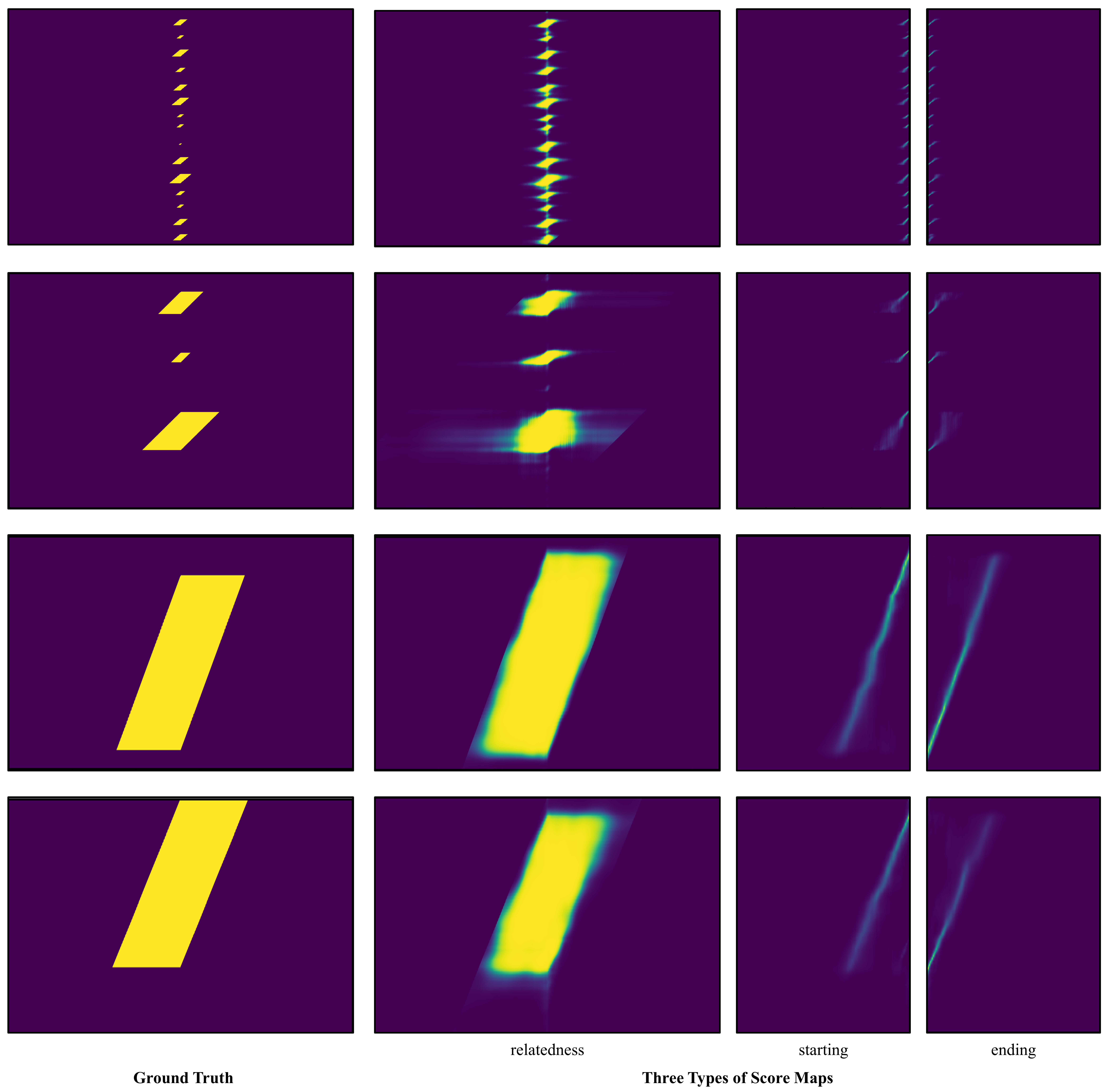}
\caption{\label{fig10}Visualization of the three types of score maps generated on THUMOS-14 (first and second rows) and ActivityNet-1.3 (third and fourth rows). We visualize the ground truth for the relatedness score map, where the left and right halves are the regions considered for the starting and ending score maps.}
\end{center}
\end{figure*}

\subsection{Ablation Study}
We conducted comprehensive ablation experiments to investigate the effectiveness of individual components of SRG.
We first performed a qualitative evaluation of the three types of score maps (i.e., relatedness, starting, and ending) produced by TIGN.
Note that the quality of these score maps is important for the generation of temporal intervals with both high recall and accurate boundaries.
We visualize these score maps as obtained on both THUMOS-14 and ActivityNet-1.3 in Fig. \ref{fig10}, where the score maps were reliably evaluated for both short and long action instances.

Second, we measured the recall with varying tIoU values for the temporal intervals obtained from the relatedness and weighted relatedness score sequences.
As shown in Table \ref{tab3}, high recall and accurate boundaries were achieved for both types of temporal intervals.
We also confirmed that incorporating both types of temporal intervals enables further improvements over the individual performances.
These temporal intervals can lead to high-quality proposals with high recall, accurate boundaries, and low false positive rates.
\begin{table}[t!]
\centering
\caption{\label{tab3}Recall@tIoU for temporal intervals generated from relatedness and weighted relatedness score sequences on THUMOS-14. RS: relatedness score sequence; WRS: weighted relatedness score sequence.}
\begin{tabular}{C{1.05cm} C{1.05cm}|C{.65cm} C{.65cm} C{.65cm} C{.65cm} C{.65cm}}\hline 
RS & WRS & 0.5 & 0.6 & 0.7 & 0.8 & 0.9 \\ \hline\hline
\checkmark &  & 93.78 & 90.22 & 83.74 & 71.28 & 49.26 \\
           & 	\checkmark & 93.75 & 89.59 & 83.71 & 70.26 & 46.28 \\
\checkmark & \checkmark & \bf{95.31} & \bf{92.16} & \bf{87.50} & \bf{75.64} & \bf{53.69} \\ \hline
\end{tabular}
\end{table}

To study the effectiveness of TIGN, TIEN, and the PN block, we measured the performance on six cases as follows: \\
\textbf{$\text{TIGN}_{\text{CM}}$:}
In TIGN, we replaced the PN block with a block consisting of two pairs of convolutional and max pooling layers, abbreviated as CM, which is identical to the early part of the PN block (see Fig. \ref{fig6}).
We used the temporal intervals generated by this TIGN as proposals.
For the action confidence evaluation, we jointly trained actionness scores and then averaged the scores of the snippets in each proposal. \\
\textbf{$\text{TIGN}_{\text{PN}}$:}
We used the temporal intervals generated by TIGN with the PN block as proposals. 
Same as $\text{TIGN}_{\text{CM}}$, we jointly trained actionness scores and averaged the scores to obtain the confidence score. \\
\textbf{$\text{TIGN}_{\text{CM}}$ + $\text{TIEN}_{\text{CM}}$:}
We generated temporal intervals by using TIGN with the CM block.
To obtain temporal action proposals, we then evaluated the action confidence scores for the temporal intervals and adjusted their boundaries by using TIEN with the CM block. \\
\textbf{$\text{TIGN}_{\text{PN}}$ + $\text{TIEN}_{\text{CM}}$:}
We produced temporal intervals by using TIGN with the PN block.
Next, we acquired proposals by using TIEN with the CM block that evaluates the confidence scores for the temporal intervals and refines their boundaries. \\
\textbf{$\text{TIGN}_{\text{CM}}$ + $\text{TIEN}_{\text{PN}}$:}
We used TIGN with the CM block to generate temporal intervals.
Then, TIEN with the PN block was used to evaluate the confidence scores for the temporal intervals and refine their boundaries to generate proposals. \\
\textbf{$\text{TIGN}_{\text{PN}}$ + $\text{TIEN}_{\text{PN}}$:}
We used TIGN with the PN block to generate temporal intervals.
We then estimated the confidence scores for the temporal intervals and adjusted their boundaries by using TIEN with the PN block for generating proposals. 
This is the proposed method, namely SRG. \\
As shown in Table \ref{tab4}, \text{$\text{TIGN}_{\text{PN}}$} achieved better performance than \text{$\text{TIGN}_{\text{CM}}$}, which verifies the effectiveness of using the PN block in TIGN.
Moreover, we achieved a significant performance gain by employing TIEN, which demonstrates that action confidence scores evaluated by TIEN are more reliable than actionness scores.
In addition, the higher improvement of \text{$\text{TIEN}_{\text{PN}}$} than \text{$\text{TIEN}_{\text{CM}}$} also proves the effectiveness of using the PN block in TIEN.
\begin{table}[t!]
\centering
\caption{\label{tab4}Study of the effectiveness of the individual components of SRG in terms of AR@AN on THUMOS-14. CM: the block consisting of two pairs of convolutional and max pooling layers; PN: pyramid non-local block.}
\setlength{\tabcolsep}{5pt}
\begin{tabular}{C{2.9cm}|C{0.66cm} C{0.66cm} C{0.66cm} C{0.66cm} C{0.66cm}} \hline
Method & @50 & @100 & @200 & @500 & @700 \\ \hline\hline
$\text{TIGN}_{\text{CM}}$ & 25.96 & 33.66 & 40.94 & 50.50 & 54.43 \\
$\text{TIGN}_{\text{PN}}$ & 29.80 & 38.21 & 46.47 & 55.61 & 59.26 \\
$\text{TIGN}_{\text{CM}}$ + $\text{TIEN}_{\text{CM}}$ & 32.56 & 40.99 & 48.18 & 56.43 & - \\
$\text{TIGN}_{\text{PN}}$ + $\text{TIEN}_{\text{CM}}$ & 34.80 & 43.23 & 51.01 & 58.30 & - \\
$\text{TIGN}_{\text{CM}}$ + $\text{TIEN}_{\text{PN}}$ & 41.47 & 48.50 & 54.86 & 61.47 & -  \\
$\text{TIGN}_{\text{PN}}$ + $\text{TIEN}_{\text{PN}}$ & \bf{42.19} & \bf{49.72} & \bf{56.71} & \bf{63.78} & - \\ \hline
\end{tabular}
\end{table}
\begin{table}[t]
\centering
\caption{\label{tab6a}Study of the effectiveness of a strategy for incorporating snippet-level scores into interval-level scores in terms of AR@AN on THUMOS-14.}
\setlength{\tabcolsep}{5pt}
\begin{tabular}{C{2.5cm}|C{1.cm} C{1.cm} C{1.cm} C{1.cm}} \hline
Method & @50 & @100 & @200 & @500 \\ \hline\hline
SRG 		& 42.19		& 49.72	 	& 56.71 & 63.78 \\ 
SRG-Boost	& \bf{42.29}	& \bf{49.83}	& \bf{56.86} & \bf{64.16} \\ \hline
\end{tabular}
\end{table}

We performed an additional experiment to investigate the possibility of boosting performance by directly incorporating snippet-level scores into interval-level scores.
The effectiveness of this strategy has been verified in several works \cite{yang2018aaai,yang2018icmr,wu2018tits}.
In our work, the relatedness scores of snippets and the action confidence scores of temporal intervals are regarded as snippet-level scores and interval-level scores, respectively.
To combine these two types of scores, we first converted the relatedness score map into a snippet-level score sequence by averaging the scores for each temporal location.
Then, we obtained a score for a given temporal interval by averaging the snippet-level scores corresponding to the locations within that temporal interval.
By multiplying this score by the action confidence score, we acquired a new action confidence score.
In Table \ref{tab6a}, we compare the performances of SRG with that of the boosted SRG, where the strategy of combining these two different scores successfully improved the performance on THUMOS-14.

\begin{table}[t!]
\centering
\caption{\label{tab5}Comparison of temporal action detection performance in terms of mAP@tIoU on THUMOS-14.}
\begin{tabular}{C{1.4cm}|C{1.7cm}|C{.5cm} C{.5cm} C{.5cm} C{.5cm} C{.5cm}}\hline 
Proposal Generator & Classifier & 0.3  & 0.4  & 0.5  & 0.6  & 0.7\\ \hline\hline
 \multicolumn{2}{c|}{S-CNN \cite{shou2016cvpr}} & 36.3 & 28.7 & 19.0 & 10.3 & 5.3 \\\hline
SST \cite{buch2017cvpr} & \multirow{6}{*}{S-CNN-cls \cite{shou2016cvpr}} & - & - & 23.0 & - & -\\ 
TURN \cite{gao2017iccv} & & 44.1 & 34.9 & 25.6 & 14.6 & 7.7\\
BSN \cite{lin2018eccv} & & 43.1 & 36.6 & 29.4 & 22.4 & 15.0\\
CTAP \cite{gao2018eccv} & & - & - & 29.9 & - & -\\
MGG \cite{liu2019cvpr} & & 44.9 & 37.8 & 29.9 & 23.6 & 15.8 \\
SRG  & & \bf{48.3} & \bf{43.6} & \bf{36.7} & \bf{28.2} & \bf{16.8} \\ \hline
SST \cite{buch2017cvpr} & \multirow{5}{*}{UNet \cite{wang2017cvpr}} & 41.2 & 31.5 & 20.0 & 10.9 & 4.7 \\ 
TURN \cite{gao2017iccv} & & 46.3 & 35.3 & 24.5 & 14.1 & 6.3\\
BSN \cite{lin2018eccv} & & 53.5 & 45.0 & 36.9 & 28.4 & 20.0\\
MGG \cite{liu2019cvpr} & & 53.9 & 46.8 & 37.4 & 29.5 & 21.3\\
SRG  & & \bf{54.5} & \bf{46.9} & \bf{39.1} & \bf{31.4} & \bf{22.2} \\ \hline
\end{tabular}
\end{table}

\begin{table}[t!]
\caption{\label{tab6}Comparison of temporal action detection performance in terms of mAP@tIoU on ActivityNet-1.3.}
\begin{tabular}{C{2.1cm}|C{1.1cm} C{1.1cm} C{1.1cm} | C{1.1cm}}\hline 
Method & 0.50 & 0.75 & 0.95 & Avg. \\ \hline\hline
TAL-Net \cite{chao2018cvpr} & 38.23 & 18.30 & 1.30 & 20.22 \\
CDC \cite{shou2017cvpr} & 45.30 & 26.00 & 0.20 & 23.80 \\
BSN \cite{lin2018eccv} + \cite{xiong2016cvprw}* & 46.45 & 29.96 & \bf{8.02} & \bf{30.03} \\
SRG + \cite{xiong2016cvprw} & \bf{46.53} & \bf{29.98} & 4.83 & 29.72 \\ \hline
\end{tabular}
\footnotesize{
*We refer to the BSN challenge paper \cite{lin2018cvprw} for the corresponding performance due to the issue reported on the performance in \cite{lin2018eccv}.}
\end{table}

\subsection{Temporal Action Detection}
To examine the quality of our proposals, we employed S-CNN-cls and UNet \cite{wang2017cvpr} as action classifiers.
S-CNN-cls is the action classifier of S-CNN \cite{shou2016cvpr}.
Specifically, we fed our proposals into S-CNN-cls and UNet to evaluate their performance for temporal action detection.
We also compared the performances of state-of-the-art proposals with the same classifiers to ours.
As the evaluation metric, we computed mean average precision (mAP) with varying tIoU values.
As shown in Table \ref{tab5}, SRG outperformed the state-of-the-art proposal generators by a large margin on the THUMOS-14 dataset.
These results demonstrate the effectiveness and superiority of SRG.

In addition, we performed a comparison of temporal action detection performance on ActivityNet-1.3 in Table \ref{tab6}, where
the performance of SRG is comparable to that of the best state-of-the-art method.
We used \cite{xiong2016cvprw} to classify the actions of our proposals.

\section{Discussion}
\begin{figure}[t!]
\centering{\includegraphics[width=.99\linewidth]{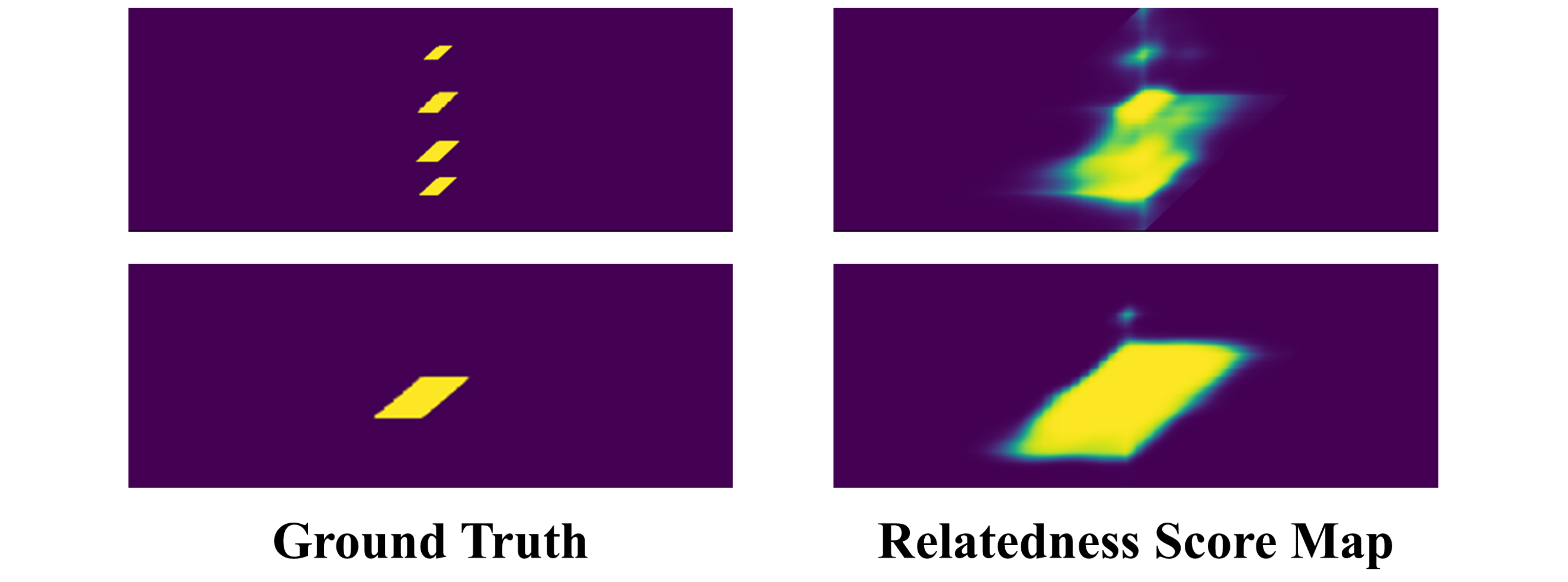}}\\
\caption{\label{fig11}Failure cases of relatedness score maps on ActivityNet-1.3.}
\end{figure}

\begin{figure}[t!]
\centering{\includegraphics[width=.99\linewidth]{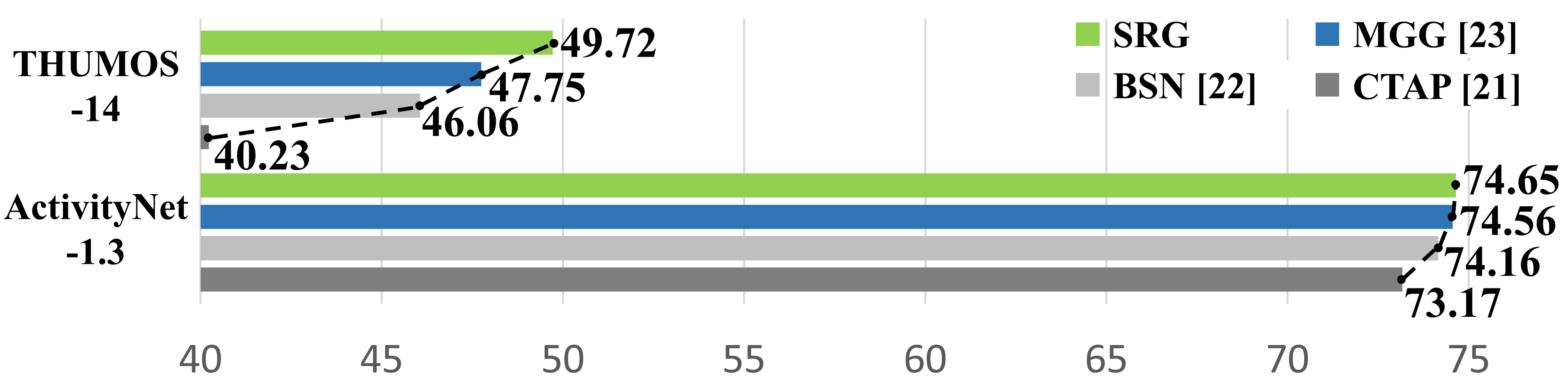}}\\
\caption{\label{fig12}Performance analysis on THUMOS-14 and ActivityNet-1.3 in terms of AR@AN=100.}
\end{figure}

In this section, we discuss the limitations of the proposed method.
In Fig. \ref{fig11}, we show failure cases of relatedness score maps for some short action instances in ActivityNet-1.3 \cite{heilbron2015cvpr}.
This is because most of the videos in ActivityNet-1.3 contain a single long instance; thus, TIGN may produce less accurate results on short action instances than long action instances.
These results can be refined by the boundary refinement of TIEN.
We further analyze the ActivityNet-1.3 and THUMOS-14 \cite{jiang2015url} datasets.
Specifically, the videos in the ActivityNet-1.3 training set contain only 1.5 action instances around the center and 36\% background per video, on average, while the videos in THUMOS-14 contain an average of 15.8 instances and 71\% background per video \cite{chao2018cvpr}.
These characteristics that often appear in real-world videos make the THUMOS-14 dataset more challenging.
We also observed marginal performance increases at a high level of AR on ActivityNet-1.3 and significant improvements at a low level of AR on THUMOS-14 (see Fig. \ref{fig12}).

Although SRG achieved the best performances at low and middle tIoU values in Fig. \ref{fig8}, there is still room for improvement at high tIoU values larger than 0.80.
Specifically, TIGN in SRG aims to generate as many temporal intervals with reliable boundaries as possible for avoiding missing action instances.
Thus, the generated temporal intervals have similar boundaries that are highly overlapped with true action instances.
Since the boundary of action is subjective and ambiguous, a large number of the highly overlapped temporal intervals make it difficult for TIEN to precisely rank them.
For example, it is a challenging task to find the best temporal intervals when they all have high tIoU values.
In a similar vein, we can explain why SRG achieved lower performance than BSN \cite{lin2018eccv} at the tIoU value of 0.95 and outperformed other methods at tIoU values of 0.50 and 0.75, as shown in Table \ref{tab6}.
For future work, we intend to address this limitation of SRG.

In Table II, SRG achieves the best performance in terms of AR@AN=100, but SRG shows comparable performance in terms of the AUC.
As aforementioned, TIGN generates a large number of temporal intervals with high tIoU values.
However, TIEN struggles to precisely rank these temporal intervals when most of the temporal intervals have good-enough boundaries.
This limitation of SRG yields low AR values at small AN values; thus, the AUC of SRG is slightly inferior to those of BSN and MGG \cite{liu2019cvpr}.
Additionally, inaccurate relatedness score maps for some short action instances on ActivityNet-1.3 (see Fig. \ref{fig11}) would limit the performance increase.
Improving these relatedness score maps will be another future work.

\section{Conclusion}
In this paper, we have proposed a novel approach, namely SRG, for high-quality temporal action proposal generation.
SRG is based on two key ideas, ``snippet relatedness'' and ``PN operations''.
The snippet relatedness indicates which snippets are related to a specific action instance.
PN operations allow the snippet relatedness to be effectively learned by locally and globally capturing long-range dependencies among snippets.
By employing these ideas, SRG first produces three types of 2D score maps for temporal intervals.
Then, SRG evaluates the action confidence scores of the temporal intervals and refines their boundaries to generate high-quality proposals.
Experimentally, we have demonstrated that SRG outperforms state-of-the-art methods on both the THUMOS-14 and ActivityNet-1.3 datasets.
Furthermore, with the same standard action classifier, our proposals markedly outperform state-of-the-art proposals for temporal action detection.


\ifCLASSOPTIONcaptionsoff
  \newpage
\fi



\bibliographystyle{IEEEtran}
\bibliography{bare_jrnl}

\begin{thebibliography}{10}
\providecommand{\url}[1]{#1}
\csname url@samestyle\endcsname
\providecommand{\newblock}{\relax}
\providecommand{\bibinfo}[2]{#2}
\providecommand{\BIBentrySTDinterwordspacing}{\spaceskip=0pt\relax}
\providecommand{\BIBentryALTinterwordstretchfactor}{4}
\providecommand{\BIBentryALTinterwordspacing}{\spaceskip=\fontdimen2\font plus
\BIBentryALTinterwordstretchfactor\fontdimen3\font minus
  \fontdimen4\font\relax}
\providecommand{\BIBforeignlanguage}[2]{{%
\expandafter\ifx\csname l@#1\endcsname\relax
\typeout{** WARNING: IEEEtran.bst: No hyphenation pattern has been}%
\typeout{** loaded for the language `#1'. Using the pattern for}%
\typeout{** the default language instead.}%
\else
\language=\csname l@#1\endcsname
\fi
#2}}
\providecommand{\BIBdecl}{\relax}
\BIBdecl

\bibitem{iwashita2013bmvc}
Y.~Iwashita, M.~Ryoo, T.~J. Fuchs, and C.~Padgett, ``Recognizing humans in
  motion: Trajectory-based aerial video analysis,'' in \emph{Proc. Brit. Mach.
  Vis. Conf.}, Sep. 2013, pp. 127.1--127.11.

\bibitem{shu2015cvpr}
T.~Shu, D.~Xie, B.~Rothrock, S.~Todorovic, and S.~C. Zhu, ``Joint inference of
  groups, events and human roles in aerial videos,'' in \emph{Proc. IEEE Conf.
  Comput. Vis. Pattern Recognit.}, Jun. 2015, pp. 4756--4584.

\bibitem{yao2015iccv}
L.~Yao, A.~Torabi, K.~Cho, N.~Ballas, C.~Pal, H.~Larocheele, and A.~Courville,
  ``Describing videos by exploiting temporal structure,'' in \emph{Proc. IEEE
  Int. Conf. Comput. Vis.}, Dec. 2015, pp. 199--211.

\bibitem{yao2016cvpr}
T.~Yao, T.~Mei, and Y.~Rui, ``Highlight detection with pairwise deep ranking
  for first-person video summarization,'' in \emph{Proc. IEEE Conf. Comput.
  Vis. Pattern Recognit.}, Jun. 2016, pp. 982--990.

\bibitem{zhang2016eccv}
K.~Zhang, W.-L. Chao, and K.~Grauman, ``Video summarization with long
  short-term memory,'' in \emph{Proc. Eur. Conf. Comput. Vis.}, Oct. 2016, pp.
  766--782.

\bibitem{chou2015tmm}
C.-L. Chou, H.-T. Chen, and S.-Y. Lee, ``Pattern-based near-duplicate video
  retrieval and localization on web-scale videos,'' \emph{IEEE Trans.
  Multimedia}, vol.~17, no.~3, pp. 382--395, Jan. 2015.

\bibitem{douze2016ijcv}
M.~Douze, J.~Revaud, J.~Varbeek, H.~J{\`e}gou, and C.~Schmid, ``Circulant
  temporal encoding for video retrieval and temporal alignment,'' \emph{Int. J.
  Comput. Vis.}, vol. 119, no.~3, pp. 291--306, Sep. 2016.

\bibitem{song2018pr}
J.~Song, L.~Gao, L.~Liu, X.~Zhu, and N.~Sebe, ``Quantization-based hashing: A
  general framework for scalable image and video retriveal,'' \emph{Pattern
  Recognit.}, vol.~75, pp. 175--187, Mar. 2018.

\bibitem{girshick2014cvpr}
R.~Girshick, J.~Donahue, T.~Darrell, and J.~Malik, ``Rich feature hierarchies
  for accurate object detection and semantic segmentation,'' in \emph{Proc.
  IEEE Conf. Comput. Vis. Pattern Recognit.}, Jun. 2014, pp. 580--587.

\bibitem{lin2017iccv}
T.-Y. Lin, P.~Goyal, R.~Girshick, K.~He, and P.~Doll{\`a}r, ``Focal loss for
  dense object detection,'' in \emph{Proc. IEEE Int. Conf. Comput. Vis.}, Oct.
  2017, pp. 2980--2988.

\bibitem{ren2015nips}
S.~Ren, K.~He, R.~Girchick, and J.~Sun, ``Faster r-cnn: Towards real-time
  object detection with region proposal networks,'' in \emph{Proc. Adv. Neural
  Inf. Process. Syst.}, Dec. 2015, pp. 91--99.

\bibitem{gao2017iccv}
J.~Gao, Z.~Yang, C.~Sun, K.~Chen, and R.~Nevatia, ``Turn tap: Temporal unit
  regression network for temporal action proposals,'' in \emph{Proc. IEEE Int.
  Conf. Comput. Vis.}, Oct. 2017, pp. 3648--3656.

\bibitem{helibron2016cvpr}
F.~C. Heilbron, J.~C. Niebles, and B.~Ghanem, ``Fast temporal activity
  proposals for efficient detection of human actions in untrimmed videos,'' in
  \emph{Proc. IEEE Conf. Comput. Vis. Pattern Recognit.}, Jun. 2016, pp.
  1914--1923.

\bibitem{shou2016cvpr}
Z.~Shou, D.~Wang, and S.-F. Chang, ``Temporal action localization in untrimmed
  videos via multi-stage cnns,'' in \emph{Proc. IEEE Conf. Comput. Vis. Pattern
  Recognit.}, Jun 2016, pp. 1049--1058.

\bibitem{chao2018cvpr}
Y.-W. Chao, S.~Vijayanarasimhan, B.~Seybold, D.~A. Ross, J.~Deng, and
  R.~Sukthankar, ``Rethinking the faster r-cnn architecture for temporal action
  localization,'' in \emph{Proc. IEEE Conf. Comput. Vis. Pattern Recognit.},
  Jun. 2018, pp. 1130--1139.

\bibitem{helibron2017cvpr}
F.~C. Heilbron, W.~Barrios, V.~Escorcia, and B.~Ghanem, ``Scc: Semantic context
  cascade for efficient action detection,'' in \emph{Proc. IEEE Conf. Comput.
  Vis. Pattern Recognit.}, Jul. 2017, pp. 3175--3184.

\bibitem{yuan2017cvpr}
Z.~Yuan, J.~C. Stroud, T.~Lu, and J.~Deng, ``Temporal action localization by
  structured maximal sums,'' in \emph{Proc. IEEE Conf. Comput. Vis. Pattern
  Recognit.}, Jul. 2017, pp. 3684--3692.

\bibitem{zhao2017iccv}
Y.~Zhao, Y.~Xiong, L.~Wang, Z.~Wu, X.~Tang, and D.~Lin, ``Temporal action
  detection with structured segment networks,'' in \emph{Proc. IEEE Int. Conf.
  Comput. Vis.}, Oct. 2017, pp. 2914--2923.

\bibitem{buch2017cvpr}
S.~Buch, V.~Escorcia, C.~Shen, B.~Ghanem, and J.~C. Niebles, ``Sst:
  Single-stream temporal action proposals,'' in \emph{Proc. IEEE Conf. Comput.
  Vis. Pattern Recognit.}, Jul. 2017, pp. 6373--6382.

\bibitem{escorcia2016eccv}
V.~Escorcia, F.~C. Heilbron, J.~C. Niebles, and B.~Ghanem, ``Daps: Deep action
  proposals for action understanding,'' in \emph{Proc. Eur. Conf. Comput.
  Vis.}, Oct. 2016, pp. 768--784.

\bibitem{gao2018eccv}
J.~Gao, K.~Chen, and R.~Nevatia, ``Ctap: Complementary temporal action proposal
  generation,'' in \emph{Proc. Eur. Conf. Comput. Vis.}, Sep. 2018, pp. 68--83.

\bibitem{lin2018eccv}
T.~Lin, X.~Zhao, H.~Su, C.~Wang, and M.~Yang, ``Bsn: Boundary sensitive network
  for temporal action proposal generation,'' in \emph{Proc. Eur. Conf. Comput.
  Vis.}, Sep. 2018, pp. 3--19.

\bibitem{liu2019cvpr}
Y.~Liu, L.~Ma, Y.~Zhang, W.~Liu, and S.-F. Chang, ``Multi-granularity generator
  for temporal action proposal,'' in \emph{Proc. IEEE Conf. Comput. Vis.
  Pattern Recognit.}, Jun. 2019, pp. 3604--3613.

\bibitem{wu2018tcsvt}
Q.~Wu, H.~Li, K.~N. Ngan, and K.~Ma, ``Image quality assessment using local
  consistency aware retriever and uncertainty aware evaluator,'' \emph{IEEE
  Trans. Circuits Syst. Video Technol.}, vol.~28, no.~9, pp. 2078--2089, Sep.
  2018.

\bibitem{lai2018eccv}
W.-S. Lai, J.-B. Huang, O.~Wang, E.~Shechtman, E.~Yumer, and M.-H. Yang,
  ``Learning blind video temporal consistency,'' in \emph{Proc. Eur. Conf.
  Comput. Vis.}, Sep. 2018, pp. 179--195.

\bibitem{dechter1997tcs}
R.~Dechter and P.~van Beek, ``Local and global relational consistency,''
  \emph{Theor. Comput. Sci.}, vol. 173, no.~1, pp. 283--308, Feb. 1997.

\bibitem{jiang2015url}
Y.~G. Jiang, J.~Liu, A.~R. Zamir, G.~Toderici, I.~Laptev, M.~Shah, and
  R.~Sukthankar, ``Thumos challenge: Action recognition with a large number of
  classes,'' 2014, http://crcv.ucf.edu/THUMOS14/.

\bibitem{heilbron2015cvpr}
F.~C. Heilbron, B.~G. V.~Escorcia, and J.~C. Niebles, ``Activitynet: A
  large-scale video benchmark for human activity understanding,'' in
  \emph{Proc. IEEE Conf. Comput. Vis. Pattern Recognit.}, Jun. 2015, pp.
  961--970.

\bibitem{efros2003iccv}
A.~A. Efros, A.~C. Berg, G.~Mori, and J.~Malik, ``Recognizing action at a
  distance,'' in \emph{Proc. IEEE Int. Conf. Comput. Vis.}, Oct. 2003, pp.
  726--733.

\bibitem{jia2008cvpr}
K.~Jia and D.-Y. Yeung, ``Human action recognition using local spatio-temporal
  discriminant embedding,'' in \emph{Proc. IEEE Conf. Comput. Vis. Pattern
  Recognit.}, Jun. 2008, pp. 1--8.

\bibitem{wang2011cvpr}
H.~Wang, A.~Kl{\"a}ser, C.~Shmid, and C.~L. Liu, ``Action recognition by dense
  trajectories,'' in \emph{Proc. IEEE Conf. Comput. Vis. Pattern Recognit.},
  Jun. 2011, pp. 3169--3176.

\bibitem{carreira2017cvpr}
J.~Carreira and A.~Zisserman, ``Quo vaids, action recognition? a new model and
  the kinetics dataset,'' in \emph{Proc. IEEE Conf. Comput. Vis. Pattern
  Recognit.}, Jul. 2017, pp. 4724--4733.

\bibitem{feichtenhofer2016cvpr}
C.~Feichtenhofer, A.~Pinz, and A.~Zisserman, ``Convolutional two-stream network
  fusion for video action recognition,'' in \emph{Proc. IEEE Conf. Comput. Vis.
  Pattern Recognit.}, Jun. 2016, pp. 1933--1941.

\bibitem{ng2015cvpr}
J.~Y.-H. Ng, M.~Hausknecht, S.~Vijayanarasimhan, O.~Vinyals, R.~Monga, and
  G.~Toderici, ``Beyond short snippets: Deep networks for video
  classificaion,'' in \emph{Proc. IEEE Conf. Comput. Vis. Pattern Recognit.},
  Jun. 2015, pp. 4694--4702.

\bibitem{simonyan2014nips}
K.~Simonyan and A.~Zisserman, ``Two-stream convolutional networks for action
  recognition in videos,'' in \emph{Proc. Adv. Neural Inf. Process. Syst.},
  Dec. 2014, pp. 568--576.

\bibitem{wang2016eccv}
L.~Wang, Y.~Xiong, Z.~Wang, Y.~Q.~D. Lin, X.~Tang, and L.~van Gool, ``Temporal
  segment networks: Towards good practices for deep action recognition,'' in
  \emph{Proc. Eur. Conf. Comput. Vis.}, Oct. 2016, pp. 20--36.

\bibitem{hara2018cvpr}
K.~Hara, H.~Kataoka, and Y.~Satoh, ``Can spatiotemporal 3d cnns retrace the
  history of 2d cnns and imagenet?'' in \emph{Proc. IEEE Conf. Comput. Vis.
  Pattern Recognit.}, Jun. 2018, pp. 6546--6555.

\bibitem{tran2015iccv}
D.~Tran, L.~Bourdev, R.~Fergus, L.~Torresani, and M.~Paluri, ``Learning
  spatiotemporal features with 3d convolutional networks,'' in \emph{Proc. IEEE
  Int. Conf. Comput. Vis.}, Dec. 2015, pp. 4489--4497.

\bibitem{xu2017iccv}
H.~Xu, A.~Das, and K.~Saenko, ``R-c3d: Region convolutional 3d network for
  temporal activity detection,'' in \emph{Proc. IEEE Int. Conf. Comput. Vis.},
  Oct. 2017, pp. 5783--5792.

\bibitem{shou2017cvpr}
Z.~Shou, J.~Chan, A.~Zareian, K.~Miyazawa, and S.-F. Chang, ``Cdc:
  Convolutional-de-convolutional networks for precise temporal action
  localization in untrimmed videos,'' in \emph{Proc. IEEE Conf. Comput. Vis.
  Pattern Recognit.}, Jul. 2017, pp. 5734--5743.

\bibitem{gao2017bmvc}
J.~Gao, Z.~Yang, and R.~Nevatia, ``Cascaded boundary regression for temporal
  action detection,'' in \emph{Proc. Brit. Mach. Vis. Conf.}, Sep. 2017, pp.
  52.1--52.11.

\bibitem{roerdink2000fi}
J.~B. Roerdink and A.~Meijster, ``The watershed transform: Definitions,
  algorithms and parallelization strategies,'' \emph{Fund. Informaticae},
  vol.~41, pp. 187--228, Apr. 2000.

\bibitem{hu2018cvpr}
J.~Hu, L.~Shen, and G.~Sun, ``Squeeze-and-excitation networks,'' in \emph{Proc.
  IEEE Conf. Comput. Vis. Pattern Recognit.}, Jun. 2018, pp. 7132--7141.

\bibitem{woo2018eccv}
S.~Woo, J.~Park, J.-Y. Lee, and I.~S. Kweon, ``Cbam: Convolutional block
  attention module,'' in \emph{Proc. Eur. Conf. Comput. Vis.}, Sep. 2018, pp.
  3--19.

\bibitem{wang2018cvpr}
X.~Wang, R.~Girshick, A.~Gupta, and K.~He, ``Non-local neural networks,'' in
  \emph{Proc. IEEE Conf. Comput. Vis. Pattern Recognit.}, Jun. 2018, pp.
  7794--7803.

\bibitem{zhao2017cvpr}
H.~Zhao, J.~Shi, X.~Qi, X.~Wang, and J.~Jia, ``Pyramid scene parsing network,''
  in \emph{Proc. IEEE Conf. Comput. Vis. Pattern Recognit.}, Jul. 2017, pp.
  2881--2890.

\bibitem{szegedy2016cvpr}
C.~Szegedy, V.~Vanhoucke, S.~Ioffe, J.~Shlens, and Z.~Wojna, ``Rethinking the
  inception architecture for computer vision,'' in \emph{Proc. IEEE Conf.
  Comput. Vis. Pattern Recognit.}, Jun. 2016, pp. 2818--2826.

\bibitem{lin2017acmmm}
T.~Lin, X.~Zhao, and Z.~Shou, ``Single shot temporal action detection,'' in
  \emph{Proc. ACM Multimedia}, Oct. 2017, pp. 988--996.

\bibitem{yang2018aaai}
K.~Yang, P.~Qiao, D.~Li, S.~Lv, and Y.~Dou, ``Exploring temporal preservation
  networks for precise temporal action localization,'' in \emph{Proc. AAAI
  Conf. Artif. Intell.}, Feb. 2018, pp. 7477--7484.

\bibitem{yang2018icmr}
H.~Qiu, Y.~Zheng, H.~Ye, Y.~Lu, F.~Wang, and L.~He, ``Precise temporal action
  localization by evolving temporal proposals,'' in \emph{Proc. Int. Conf.
  Multimedia Retr.}, Jun. 2018, pp. 388--396.

\bibitem{wu2018tits}
Q.~Wu, H.~Li, F.~Meng, and K.~N. Ngan, ``Generic proposal evaluator: A lazy
  learning strategy toward blind proposal quality assessment,'' \emph{IEEE
  Trans. Intell. Transp. Syst.}, vol.~19, no.~1, pp. 306--319, Jan. 2018.

\bibitem{wang2017cvpr}
L.~Wang, Y.~Xiong, D.~Lin, and L.~van Gool, ``Untrimmednets for weakly
  supervised action recognition and detection,'' in \emph{Proc. IEEE Conf.
  Comput. Vis. Pattern Recognit.}, Jul. 2017, pp. 4325--4334.

\bibitem{xiong2016cvprw}
Y.~Xiong, L.~Wang, Z.~Wang, B.~Zhang, H.~Song, W.~Li, D.~Lin, Y.~Qiao, L.~van
  Gool, and X.~Tang, ``Cuhk \& ethz \& siat submission to activitynet challenge
  2016,'' in \emph{Proc. IEEE Conf. Comput. Vis. Pattern Recognit. Workshop},
  Jun. 2016.

\bibitem{lin2018cvprw}
T.~Lin, H.~Su, and X.~Zhao, ``Boundary sensitive network: Submission to
  activitynet challenge 2018,'' in \emph{Proc. IEEE Conf. Comput. Vis. Pattern
  Recognit. Workshop}, Jun. 2018.

\end{thebibliography}
\end{document}